\documentclass[10pt]{article}

\usepackage{opdarena-preprint}

\usepackage[utf8]{inputenc}
\usepackage[T1]{fontenc}
\usepackage{float}
\usepackage{placeins}
\usepackage{hyperref}
\usepackage{xurl}
\usepackage{booktabs}
\usepackage{amsfonts}
\usepackage{amsmath}
\usepackage{microtype}
\usepackage[table]{xcolor}
\usepackage{tikz}
\usepackage{graphicx}
\usepackage{multirow}

\newcommand{\rawresultfigure}[3]{%
  \begin{figure}[H]
  \centering
  \includegraphics[width=\textwidth,height=.82\textheight,keepaspectratio]{figures/#1}
  \caption{Task accuracies during training for #2 Base.}
  \label{fig:raw-#3}
  \end{figure}
}
\newcommand{\rawresultpage}[3]{%
  \clearpage
  \rawresultfigure{#1}{#2}{#3}
}

\title{OPDArena: Understanding Learning and\\Forgetting in On-Policy Distillation}

\newcommand{\paperauthors}{Yuanze Xu, Tianzhu Ye, Hui Sun, Wenhui Wang, Li Dong, Furu Wei}
\author[1]{Yuanze Xu\thanks{Equal contribution.\
  \textsuperscript{\textdagger}Corresponding author.}}
\author[1]{Tianzhu Ye\textsuperscript{*}}
\author[2]{Hui Sun}
\author[1]{\authorcr Wenhui Wang}
\author[1]{Li Dong}
\author[1]{Furu Wei\textsuperscript{\textdagger}}
\affil[1]{Microsoft Research}
\affil[2]{Nankai University}
\date{}

\newcommand{\papercodelink}{\href{https://github.com/StellarCurtain/OPDArena}{\nolinkurl{github.com/StellarCurtain/OPDArena}}}

\hypersetup{
    pdftitle={OPDArena: Understanding Learning and Forgetting in On-Policy Distillation},
    pdfauthor={\paperauthors},
    hidelinks
}

\begin{document}

\raggedbottom

\maketitle

\begin{abstract}
On-policy distillation (OPD) is increasingly used for language model
post-training, yet how teacher supervision shapes learning and forgetting
remains unclear. We present OPDArena, a framework comparing supervised
fine-tuning (SFT), reinforcement learning with verifiable rewards (RLVR),
and OPD across five pretrained models and diverse teacher sources.
It jointly evaluates mathematical learning, general capability retention,
and predictive distribution changes. Teacher choice strongly shapes
retention: students of RLVR-trained teachers improve on the
target task while retaining prior capabilities, whereas students of
SFT-trained teachers incur greater capability loss.
Across post-training methods, out-of-domain (OOD) distribution
change shows a more consistent positive association with forgetting than
in-domain change. To understand how teacher choice affects these changes,
we examine the additional information teachers provide relative to the
student's pretrained model. Controlled teacher comparisons show that the
task-specific preferences teachers acquire through post-training make
student updates more concentrated in-domain.
\end{abstract}

{\centering\small\textbf{Code:} \papercodelink\par}

\section{Introduction}
\label{sec:intro}

Post-training enables language models to acquire new skills, but can
erode capabilities learned during
pretraining~\citep{wang2024continualsurvey,luo2025forgetting}.
Compared with supervised fine-tuning (SFT) on fixed, off-policy responses,
reinforcement learning with verifiable rewards (RLVR) often preserves
prior capabilities better, a behavior linked to on-policy data and limited
distributional change~\citep{shenfeld2025rlrazor,zhang2026rft}.
On-policy distillation (OPD) combines student-generated responses with
token-level teacher supervision~\citep{gu2024minillm,agarwal2024onpolicy,li2026rethinkingopd}.
OPD shares the on-policy setting, yet retention outcomes vary substantially
across teacher configurations~\citep{ye2026opcd,wang2026denser}.
\textit{This raises the question of how OPD compares with SFT and RLVR
in learning and capability retention, and how teacher choice shapes
these outcomes.}

\begin{figure}[!t]
    \centering
    \includegraphics[width=0.94\textwidth]{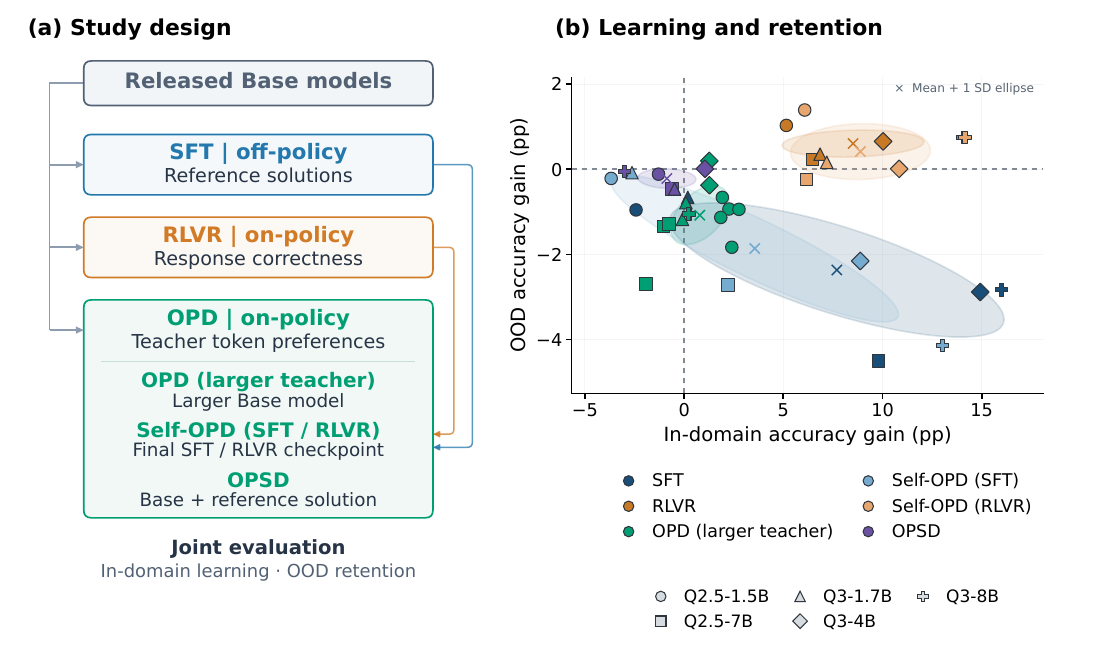}
    \caption{Training methods and joint learning--retention outcomes.
    (a) Training paths and feedback sources from
    a shared Base model. (b) In-domain accuracy gain versus OOD accuracy
    gain from Base; colors identify methods and shapes identify models.
    Ellipses summarize variation across model--teacher configurations.
    RLVR and Self-OPD (RLVR) combine task learning with low forgetting.}
    \label{fig:overview}
\end{figure}

To answer the above question, we introduce OPDArena, a framework that
systematically compares SFT, RLVR, and OPD across five Base models and
four types of OPD teachers under a shared mathematical post-training setup.
Figure~\ref{fig:overview} summarizes this comparison and reveals distinct
learning and retention patterns: 1) RLVR and Self-OPD (RLVR) combine
mathematical gains with strong retention, while on-policy self-distillation
(OPSD)~\citep{zhao2026opsd} stays close to Base accuracy in both domains.
2) OPD from larger teachers and Self-OPD (SFT) incur more severe forgetting,
while SFT has the largest average capability loss.

To understand these differences, we examine changes in predictive
distributions. Across methods, OOD change shows a more consistent positive
association with forgetting than in-domain change. This motivates
examining update selectivity, measured by OOD prediction change per unit
of in-domain change, in relation to the information teachers provide
relative to student Base.
Controlled comparisons of OPD teachers show that the preferences
they acquire through SFT or RLVR concentrate student updates more strongly
in-domain. The student's own distribution further shapes
how these preferences contribute to updates, helping explain the different
rates at which task gains and forgetting are inherited.

\Needspace{4\baselineskip}
Our contributions are:

\begin{itemize}
    \item \textbf{A systematic evaluation framework.}
    OPDArena compares SFT, RLVR, and OPD teacher sources under a shared
    setup, jointly characterizing target-domain learning, general
    capability retention, and predictive distribution changes.
    \item \textbf{A refined view of distribution change and retention.}
    We show that OOD distribution change is more consistently associated
    with forgetting than in-domain change across student models and
    teacher configurations. We further use the OOD-to-in-domain change ratio,
    $\rho$, to characterize update selectivity.
    \item \textbf{Teacher information and update selectivity.}
    Controlled comparisons of OPD teachers show that
    preferences acquired through SFT or RLVR improve student update
    selectivity. Further analyses identify mismatches between
    teacher-preferred responses and the student's distribution that can
    hinder task learning, helping explain why task gains and forgetting
    transfer at different rates.
\end{itemize}

\section{Method}
\label{sec:method}

We compare post-training methods applied to the same released Qwen Base models
and measure their effects on mathematical learning and general capability
retention. Our framework distinguishes how supervision enters the model
from how its predictions change across domains.
We first describe the training signals and teacher
configurations (Section~\ref{sec:training_signals}), then define how we measure
learning, forgetting, and distribution change (Section~\ref{sec:distribution_measures}).

\subsection{Training Signals and Teachers}
\label{sec:training_signals}

Let $p_0$ denote a Base model; $p_\theta$ is the model being trained
from it, and $q$ is a frozen distillation teacher.
The training set $\mathcal D$ contains problems $x$ and reference
solutions $c$. SFT learns from these fixed, off-policy solutions;
RLVR and OPD use on-policy responses $y$ generated by the model
being trained. At response-token position $t\in\{1,\ldots,|y|\}$,
where $|y|$ is the response length, the student prefix is
$s_t=(x,y_{<t})$; $s_t^{\mathrm T}$ denotes the same problem and
continuation with the teacher's conditioning information.
The training objectives are:
\begin{equation}
    \mathcal L_{\mathrm{SFT}}(\theta)
    =-\mathbb E_{(x,c)\sim\mathcal D}
      \left[\sum_t\log p_\theta(c_t\mid x,c_{<t})\right],
    \label{eq:sft_objective}
\end{equation}
\begin{equation}
    \mathcal L_{\mathrm{RLVR}}(\theta)
    =-\mathbb E_{x\sim\mathcal D,\;y\sim p_\theta(\cdot\mid x)}
      \left[R(x,y)\right],
    \label{eq:rlvr_objective}
\end{equation}
\begin{equation}
    \mathcal L_{\mathrm{OPD}}(\theta;\bar\theta)
    =\mathbb E_{s\sim d_{\bar\theta}}
      \left[\mathrm{KL}\bigl(p_\theta(\cdot\mid s)
      \,\|\,q(\cdot\mid s^{\mathrm T})\bigr)\right].
    \label{eq:opd_objective}
\end{equation}
In Eq.~\eqref{eq:rlvr_objective}, $R(x,y)$ is a verifiable correctness reward.
In Eq.~\eqref{eq:opd_objective}, $d_{\bar\theta}$ is the empirical prefix
distribution collected with the current rollout policy $p_{\bar\theta}$,
held fixed during the actor update. Reverse Kullback--Leibler (KL)
divergence transfers teacher
preferences at these student-generated
prefixes~\citep{gu2024minillm,agarwal2024onpolicy}.

\paragraph{Training feedback.}
Our RLVR baseline uses group relative policy optimization
(GRPO)~\citep{shao2024deepseekmath} with verifiable
correctness rewards. Centering and scaling rewards within each group
of responses to the same problem gives a relative advantage shared
by all tokens in a response.
OPD assigns token-dependent advantages through the
teacher--student log-probability ratio. We use a student-weighted
Top-16 approximation~\citep{li2026rethinkingopd}.
Input formats and the actor update settings are
given in Appendices~\ref{app:training_input_formats}
and~\ref{app:training_settings}.

\paragraph{Teacher source.}
We study self-distillation from post-trained checkpoints, distinguishing
Self-OPD (SFT) and Self-OPD (RLVR) by how the teacher is trained.
We reuse the final saved checkpoints of the corresponding SFT and
RLVR experiments reported here as frozen teachers.
Each student starts from the same released Base model used in its
teacher's training. OPD from larger models uses frozen Base teachers with more
parameters than the student, drawn from the same or a different model family.
We use OPSD~\citep{zhao2026opsd} as a representative of self-distillation
with privileged information, where the teacher receives additional
context unavailable to the student. Within our shared OPD recipe, a frozen
copy of the student's Base model receives the problem and a reference solution, while
the student receives only the problem and learns from the teacher's
probabilities on student-generated prefixes.

\subsection{Capability and Distribution Measurements}
\label{sec:distribution_measures}

For a task group $\mathcal G$, let $\operatorname{Acc}_{\mathcal G}(p)$
denote model $p$'s accuracy averaged equally across tasks in $\mathcal G$.
We measure capability changes relative to Base:
\begin{equation}
    \Delta\operatorname{Acc}_{d}(p_\theta)
    =\operatorname{Acc}_{d}(p_\theta)
    -\operatorname{Acc}_{d}(p_0),
    \qquad d\in\{\mathrm{in},\mathrm{OOD}\}.
    \label{eq:accuracy_change}
\end{equation}
Positive in-domain changes indicate learning gains, while negative OOD
changes indicate forgetting. All changes are expressed in percentage points.

To compare the predictive distributions of any two models $p$ and $q$,
we use Jensen--Shannon (JS) divergence for a symmetric, bounded comparison.
We evaluate their next-token predictions on the same fixed
prefixes, with $\nu_j$ denoting the prefix distribution for task $j$:
\begin{equation}
    D_{\mathcal G}(p,q)
    =\frac{1}{|\mathcal G|}\sum_{j\in\mathcal G}
    \mathbb E_{s\sim\nu_j}
    \left[\operatorname{JS}\bigl(p(\cdot\mid s),q(\cdot\mid s)\bigr)\right].
    \label{eq:distribution_change}
\end{equation}
We use natural logarithms and
$\operatorname{JS}(p,q)=\tfrac12\mathrm{KL}(p\|m)+\tfrac12\mathrm{KL}(q\|m)$,
where $m=(p+q)/2$. Distribution measurements use the full vocabulary
and weight datasets equally. Prefix sampling is described in
Appendix~\ref{app:distribution_measurements}.

\section{Experiments}
\label{sec:experiments}

\subsection{Experimental Setup}
\label{sec:experimental_setup}

\paragraph{Models and methods.}
We experiment with the Qwen2.5~\citep{yang2024qwen25} (1.5B and 7B)
and Qwen3~\citep{yang2025qwen3} (1.7B, 4B, and 8B) model families.
We use their pretrained Base versions to compare post-training methods
and teacher sources from a shared starting point before instruction tuning.
For each model, all methods start from the same Base checkpoint.
We compare the SFT, RLVR, and OPD configurations defined in
Section~\ref{sec:training_signals}.
As shown in Figure~\ref{fig:overview}(a), Self-OPD
uses an SFT or RLVR endpoint as the frozen teacher for a new student
initialized from Base.

\paragraph{Training.}
We train the five Base models on OpenR1-50K for one epoch, following
previous work~\citep{huang2026peftarena,li2026rethinkingopd}.
The models being trained receive Raw prompts without chat templates,
matching the Base models' text-completion format and keeping prompt
formatting consistent across methods. Our Raw/Template baseline
comparison supports this choice (Appendix~\ref{app:base_model_baselines}).

\paragraph{Evaluation.}
We evaluate in-domain mathematical reasoning on AMC23 and
AIME24~\citep{li2024numinamath}, AIME25 and
HMMT25~\citep{balunovic2025matharena}, and Math500~\citep{lightman2024verify}.
OOD evaluations use HellaSwag~\citep{zellers2019hellaswag},
ARC-Challenge~\citep{clark2018arc}, MMLU~\citep{hendrycks2021mmlu},
IFEval~\citep{zhou2023ifeval}, TriviaQA~\citep{joshi2017triviaqa},
BBH~\citep{suzgun2023bbh}, and NQ-Open~\citep{kwiatkowski2019natural,lee2019latent}.
For each group, we report mean accuracy, weighting the five in-domain
or seven OOD datasets equally. Main results are averaged over two training seeds.
Full training and evaluation settings,
including original-model baselines, are provided in
Appendix~\ref{app:experimental_details}.

\subsection{Main Results}
\label{sec:main_results}

Table~\ref{tab:main_results} summarizes in-domain and OOD accuracy for four
of the five Base models; Appendix~\ref{app:complete_raw} provides training
curves and individual-task results for all five.
Accuracies are averaged equally over datasets, with changes from Base
in parentheses.
Negative OOD changes indicate forgetting.

\begin{table}[!t]
    \centering
    \caption{Main results for four Base models: accuracy (\%) with changes
    from Base in parentheses (percentage points). Negative OOD changes
    indicate forgetting. RLVR and Self-OPD (RLVR) combine in-domain
    gains with strong capability retention.}
    \label{tab:main_results}
    \begingroup
    \small
    \setlength{\tabcolsep}{0pt}
    \renewcommand{\arraystretch}{1.12}
    \definecolor{resultgain}{HTML}{147D64}
    \definecolor{resultloss}{HTML}{A44535}
    \definecolor{resultneutral}{HTML}{707070}
    \newlength{\resultvaluewidth}
    \newlength{\resultchangewidth}
    \newlength{\resultcellwidth}
    \settowidth{\resultvaluewidth}{00.00}
    \settowidth{\resultchangewidth}{{\scriptsize\ensuremath{(+00.00)}}}
    \newcommand{\resultcell}[3]{\makebox[\resultvaluewidth][r]{#1}\,\makebox[\resultchangewidth][r]{{\scriptsize\textcolor{#3}{\ensuremath{(#2)}}}}}
    \settowidth{\resultcellwidth}{\resultcell{00.00}{+00.00}{resultneutral}}
    \newcommand{\basecell}[1]{\makebox[\resultvaluewidth][r]{#1}\,\makebox[\resultchangewidth][r]{}}
    \newcommand{\resultheading}[1]{\makebox[\resultcellwidth][c]{#1}}
    \begin{tabular*}{\textwidth}{@{\extracolsep{\fill}}lrrrr@{}}
        \toprule
        \textbf{Method (Teacher)} & \multicolumn{2}{c}{\textbf{Qwen2.5-1.5B}} & \multicolumn{2}{c}{\textbf{Qwen2.5-7B}} \\
        \cmidrule(l{2pt}r{2pt}){2-3} \cmidrule(l{2pt}r{2pt}){4-5}
         & \resultheading{In-domain} & \resultheading{OOD} & \resultheading{In-domain} & \resultheading{OOD} \\
        \midrule
        Base & \basecell{13.93} & \basecell{35.20} & \basecell{26.08} & \basecell{49.82} \\
        \midrule
        SFT & \resultcell{11.50}{-2.42}{resultloss} & \resultcell{34.25}{-0.95}{resultloss} & \resultcell{35.87}{+9.79}{resultgain} & \resultcell{45.32}{-4.50}{resultloss} \\
        RLVR & \resultcell{19.08}{+5.15}{resultgain} & \resultcell{36.23}{+1.03}{resultgain} & \resultcell{32.56}{+6.48}{resultgain} & \resultcell{50.06}{+0.23}{resultgain} \\
        \addlinespace[3pt]
        OPD (Q2.5-7B) & \resultcell{15.85}{+1.93}{resultgain} & \resultcell{34.54}{-0.66}{resultloss} & \basecell{--} & \basecell{--} \\
        OPD (Q2.5-14B) & \resultcell{16.18}{+2.26}{resultgain} & \resultcell{34.27}{-0.93}{resultloss} & \resultcell{25.03}{-1.05}{resultloss} & \resultcell{48.47}{-1.35}{resultloss} \\
        OPD (Q3-4B) & \resultcell{15.77}{+1.85}{resultgain} & \resultcell{34.07}{-1.13}{resultloss} & \basecell{--} & \basecell{--} \\
        OPD (Q3-8B) & \resultcell{16.33}{+2.40}{resultgain} & \resultcell{33.37}{-1.83}{resultloss} & \resultcell{24.17}{-1.91}{resultloss} & \resultcell{47.13}{-2.70}{resultloss} \\
        OPD (Q3-14B) & \resultcell{16.69}{+2.76}{resultgain} & \resultcell{34.26}{-0.94}{resultloss} & \resultcell{25.32}{-0.76}{resultloss} & \resultcell{48.54}{-1.29}{resultloss} \\
        \addlinespace[3pt]
        Self-OPD (SFT) & \resultcell{10.25}{-3.68}{resultloss} & \resultcell{34.99}{-0.21}{resultloss} & \resultcell{28.29}{+2.21}{resultgain} & \resultcell{47.11}{-2.72}{resultloss} \\
        Self-OPD (RLVR) & \resultcell{20.00}{+6.08}{resultgain} & \resultcell{36.60}{+1.39}{resultgain} & \resultcell{32.25}{+6.17}{resultgain} & \resultcell{49.59}{-0.24}{resultloss} \\
        \addlinespace[3pt]
        OPSD & \resultcell{12.64}{-1.29}{resultloss} & \resultcell{35.09}{-0.11}{resultloss} & \resultcell{25.46}{-0.62}{resultloss} & \resultcell{49.35}{-0.47}{resultloss} \\
        \bottomrule
    \end{tabular*}
    \par\vspace{8pt}
    \begin{tabular*}{\textwidth}{@{\extracolsep{\fill}}lrrrr@{}}
        \toprule
        \textbf{Method (Teacher)} & \multicolumn{2}{c}{\textbf{Qwen3-4B}} & \multicolumn{2}{c}{\textbf{Qwen3-8B}} \\
        \cmidrule(l{2pt}r{2pt}){2-3} \cmidrule(l{2pt}r{2pt}){4-5}
         & \resultheading{In-domain} & \resultheading{OOD} & \resultheading{In-domain} & \resultheading{OOD} \\
        \midrule
        Base & \basecell{26.85} & \basecell{45.45} & \basecell{31.32} & \basecell{51.37} \\
        \midrule
        SFT & \resultcell{41.77}{+14.93}{resultgain} & \resultcell{42.57}{-2.88}{resultloss} & \resultcell{47.32}{+16.00}{resultgain} & \resultcell{48.54}{-2.83}{resultloss} \\
        RLVR & \resultcell{36.87}{+10.03}{resultgain} & \resultcell{46.11}{+0.66}{resultgain} & \resultcell{45.39}{+14.07}{resultgain} & \resultcell{52.12}{+0.75}{resultgain} \\
        \addlinespace[3pt]
        OPD (Q3-8B) & \resultcell{28.12}{+1.28}{resultgain} & \resultcell{45.08}{-0.38}{resultloss} & \basecell{--} & \basecell{--} \\
        OPD (Q3-14B) & \resultcell{28.11}{+1.27}{resultgain} & \resultcell{45.65}{+0.20}{resultgain} & \resultcell{31.56}{+0.24}{resultgain} & \resultcell{50.32}{-1.05}{resultloss} \\
        \addlinespace[3pt]
        Self-OPD (SFT) & \resultcell{35.72}{+8.88}{resultgain} & \resultcell{43.30}{-2.15}{resultloss} & \resultcell{44.34}{+13.02}{resultgain} & \resultcell{47.23}{-4.14}{resultloss} \\
        Self-OPD (RLVR) & \resultcell{37.68}{+10.84}{resultgain} & \resultcell{45.46}{+0.01}{resultgain} & \resultcell{45.48}{+14.16}{resultgain} & \resultcell{52.12}{+0.75}{resultgain} \\
        \addlinespace[3pt]
        OPSD & \resultcell{27.89}{+1.05}{resultgain} & \resultcell{45.46}{+0.01}{resultgain} & \resultcell{28.31}{-3.01}{resultloss} & \resultcell{51.32}{-0.05}{resultloss} \\
        \bottomrule
    \end{tabular*}
    \endgroup
\end{table}

\paragraph{Learning and capability retention.}
As shown in Figure~\ref{fig:overview}(b),
RLVR and Self-OPD (RLVR) combine mathematical gains with OOD accuracy close to Base.
SFT and Self-OPD (SFT)
generally incur larger OOD losses.
Thus, Self-OPD with SFT and RLVR teachers shows different retention
patterns even though both variants learn from student-generated responses.
Limited SFT gains and modest forgetting in the smallest models are
consistent with prior observations of difficulty learning long reasoning
traces~\citep{li2025smallmodels} and milder forgetting at smaller
scales~\citep{luo2025forgetting}.
OPD from larger Base teachers yields modest or inconsistent mathematical
gains, accompanied by some OOD capability loss.
OPSD largely preserves OOD accuracy, with details of its adaptation to
Base models in Appendix~\ref{app:opsd_formats}.

\FloatBarrier

\section{Analysis}
\label{sec:discussion}

We study the relationship between predictive distribution change and
forgetting during post-training (Section~\ref{sec:distribution_forgetting}).
We then focus on OPD, examining how teacher information sources shape
learning and retention (Section~\ref{sec:teacher_information}) and how
students inherit teacher behavior through parameter and prediction
trajectories (Section~\ref{sec:teacher_target}).

\subsection{Distribution Change and Forgetting}
\label{sec:distribution_forgetting}
\label{sec:predictive_distributions}

Previous work links forgetting to KL divergence from the initial model
on the training task~\citep{shenfeld2025rlrazor}.
We extend this perspective by comparing in-domain and OOD prediction
changes across SFT, RLVR, and OPD.
Using Eq.~\eqref{eq:distribution_change}, we write
$D_{\mathrm{in}}=D_{\mathrm{in}}(p_0,p_t)$ and
$D_{\mathrm{OOD}}=D_{\mathrm{OOD}}(p_0,p_t)$ for prediction changes
from the pretrained Base model $p_0$ to $p_t$, the model at training step $t$.

\begin{figure}[!htb]
    \centering
    \includegraphics[width=\textwidth]{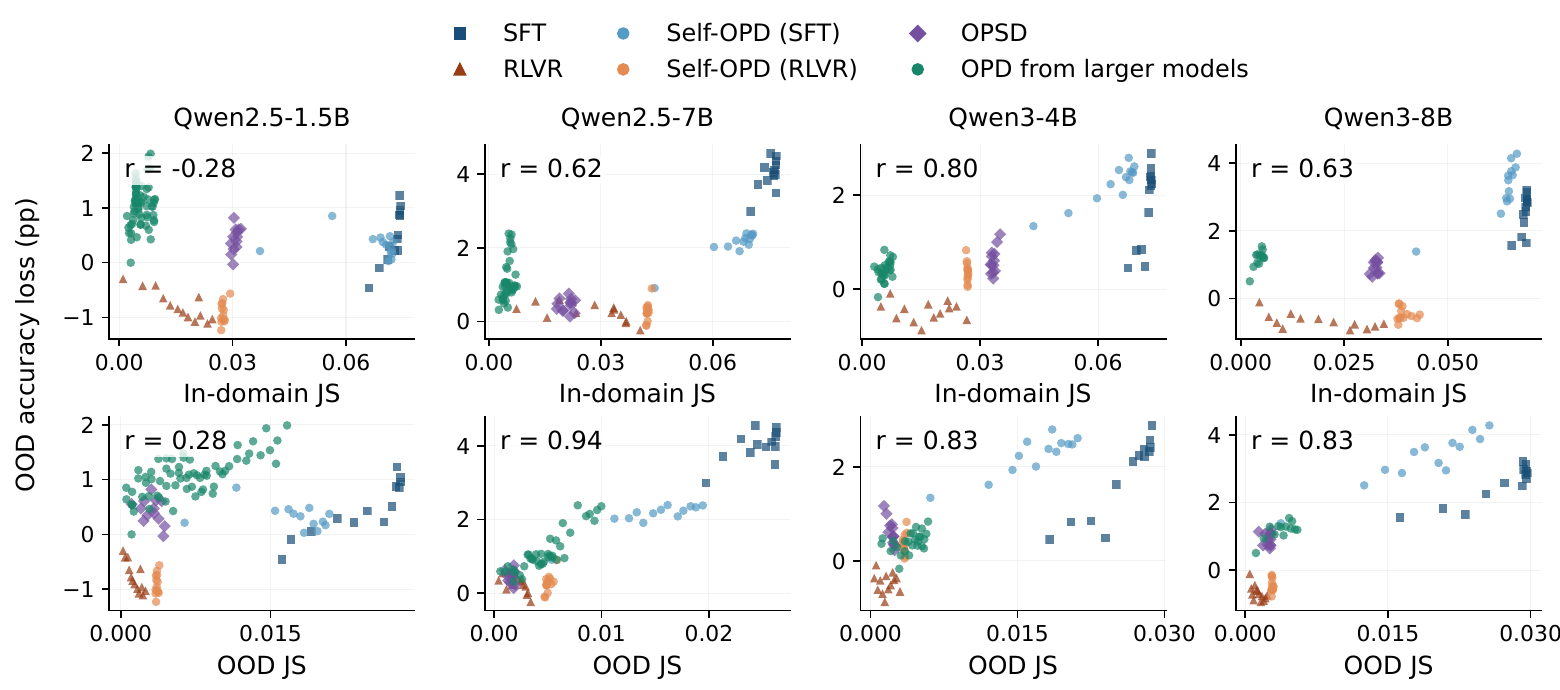}
    \caption{Distribution change and forgetting.
    Horizontal axes show in-domain (top) and OOD (bottom) JS from Base;
    larger values indicate greater distribution change. The vertical axis shows
    OOD accuracy loss from Base; higher values indicate greater forgetting.
    Each point pairs JS and OOD accuracy at the same checkpoint.
    JS is in nats, averaged equally over five in-domain or seven OOD tasks;
    $r$ is Pearson correlation.
    Qwen3-1.7B is shown in Appendix Figure~\ref{fig:distribution_forgetting_q3_1p7b}.}
    \label{fig:distribution_forgetting}
    \label{fig:forgetting_correlation_ood}
\end{figure}

\paragraph{OOD prediction change is more consistently associated with forgetting.}
Figure~\ref{fig:distribution_forgetting} compares checkpoints across
SFT, RLVR, and OPD configurations. OOD JS, which measures prediction changes
on the same tasks used to evaluate forgetting, is positively correlated
with forgetting for each model, with stronger associations for the larger models.

\paragraph{Relative OOD change measures update selectivity.}
This association motivates seeking task learning with less accompanying
OOD change. Since methods also produce different amounts of in-domain
change, we define
\begin{equation}
    \rho=\frac{D_{\mathrm{OOD}}}{D_{\mathrm{in}}},
    \qquad D_{\mathrm{in}}>0.
    \label{eq:ood_change_factors}
\end{equation}
The ratio measures OOD change per unit of in-domain prediction change;
lower $\rho$ indicates greater update selectivity, with changes more
concentrated in the target domain.

\begin{figure}[!htb]
    \centering
    \includegraphics[width=.814\textwidth]{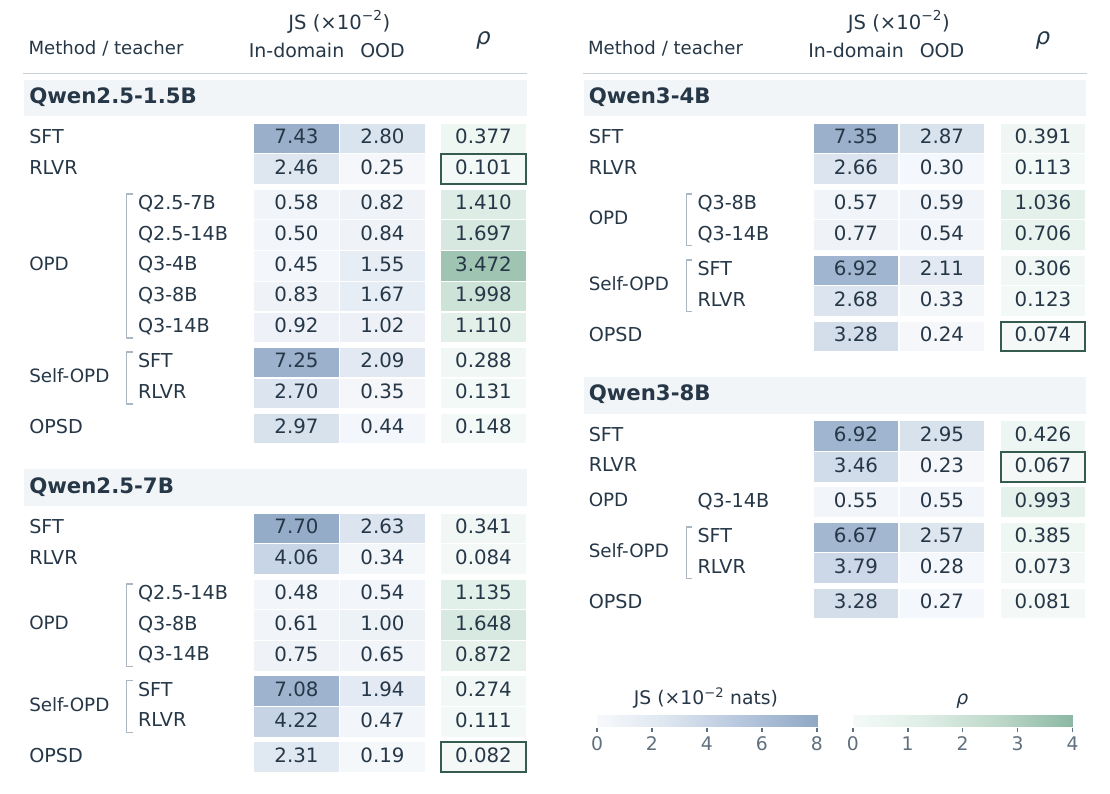}
    \caption{Prediction changes from Base across post-training methods.
    The first two columns show in-domain and OOD JS in $10^{-2}$ nats;
    the last shows relative OOD influence $\rho$.
    Boxes mark the lowest $\rho$ for each model.}
    \label{fig:rawraw_predictive_distributions}
    \label{fig:predictive_distributions}
\end{figure}

Figure~\ref{fig:predictive_distributions} applies this measure across
post-training methods. Within OPD, larger Base teachers produce small
in-domain changes and high $\rho$, while post-trained teachers and
supervision with reference solutions distribute changes differently.
These differences under a shared distillation objective motivate
examining the information provided by each teacher.

\subsection{Teacher Information Sources and Relative OOD Influence}
\label{sec:teacher_information}

Given the association between OOD prediction change and forgetting,
we examine how teacher choice affects OPD's update selectivity and
retention of OOD capabilities.

\paragraph{Stronger teacher OOD capabilities can accompany greater student forgetting.}
Table~\ref{tab:teacher_capability_counterexample} compares a Qwen3-8B Base
teacher with each student's own RLVR-trained counterpart for Qwen2.5-1.5B
and Qwen2.5-7B. Despite stronger OOD capability and comparable or better
mathematical capability, the Base teacher yields lower student mathematical
accuracy, more forgetting, and higher $\rho$ in both comparisons.

\begin{table}[!htbp]
    \centering
    \small
    \setlength{\tabcolsep}{8pt}
    \renewcommand{\arraystretch}{1.12}
    \begin{tabular}{lrrp{8pt}rrr}
\toprule
Teacher & \multicolumn{2}{c}{Teacher acc.} & & \multicolumn{2}{c}{Student acc.} &  \\
\cmidrule(lr){2-3}\cmidrule(lr){5-6}
& Math & OOD & & Math & OOD & $\rho$ \\
\midrule
\rowcolor{black!5}
\multicolumn{4}{l}{\textbf{Student: Qwen2.5-1.5B Base}} & 13.93 & 35.20 & \\
Qwen3-8B Base & 31.32 & \textbf{51.37} & & 16.33 & 33.37 & 1.998 \\
Qwen2.5-1.5B RLVR & 19.08 & 36.23 & & 20.00 & \textbf{36.60} & 0.131 \\
\midrule
\rowcolor{black!5}
\multicolumn{4}{l}{\textbf{Student: Qwen2.5-7B Base}} & 26.08 & 49.82 & \\
Qwen3-8B Base & 31.32 & \textbf{51.37} & & 24.17 & 47.13 & 1.648 \\
Qwen2.5-7B RLVR & 32.56 & 50.06 & & 32.25 & \textbf{49.59} & 0.111 \\
\bottomrule
\end{tabular}

    \caption{Teacher and student accuracies (\%) and student $\rho$ at
    step 780. Each block compares a larger Base teacher with the student's
    own RLVR teacher; shaded headings identify the student and its Base accuracies.
    Datasets are weighted equally within each domain.}
    \label{tab:teacher_capability_counterexample}
\end{table}

\paragraph{Teacher information and update selectivity.}
These OPD configurations produce markedly different retention and update
selectivity. With the student initialization and distillation setup fixed,
they differ in the additional information supplied by the teacher.
We therefore examine what each
teacher adds relative to student Base and how this information shapes
update selectivity.
We hypothesize that task-specific
preferences concentrate updates in-domain, while general response habits
also affect OOD predictions~\citep{wang2026denser}:
\begin{itemize}
    \setlength{\itemsep}{3pt}
    \setlength{\parsep}{0pt}
    \item \textbf{OPD from larger Base teachers.} These teachers produce
    high student $\rho$.
    Their feedback reflects cross-model pretraining differences spanning
    mathematics and many other tasks. Even when OPD trains only on
    mathematical problems, fitting these broader preferences may change
    OOD predictions through shared parameters, contributing to higher $\rho$.
    \item \textbf{Self-OPD (SFT).} Starting from the student's Base,
    SFT fits the teacher's predictions
    to fixed external reference answers through off-policy training,
    learning their solution content and response habits. Self-OPD then
    conveys these acquired preferences
    through the teacher's token-level predictions on student-generated
    responses. The inherited response habits may also affect OOD behavior.
    \item \textbf{Self-OPD (RLVR).} Starting from the student's Base,
    RLVR adjusts the teacher's distribution
    using correctness scores on its own on-policy responses. Self-OPD then
    conveys these reward-based adjustments through the teacher's token-level
    predictions on student-generated responses. The acquired preference for
    higher-reward behavior may concentrate feedback more strongly on the
    target task.
    \item \textbf{OPSD.} Teacher weights remain identical to student Base,
    while an in-domain mathematical reference solution supplies problem-specific evidence through
    context. This evidence may concentrate feedback on the current task.
\end{itemize}
This hypothesis predicts that adding mathematical post-training to the
same larger Base teacher will make its feedback more task-specific and
lower student $\rho$.

\paragraph{Decomposing teacher preferences.}
Let $p_0$ be student Base, $q_{\mathrm{base}}$ the larger Base, and $q_m$ its
endpoint after $m\in\{\mathrm{SFT},\mathrm{RLVR}\}$ training. The teacher's
preferences relative to student Base decompose as
\begin{equation}
    \log\frac{q_m(v\mid s)}{p_0(v\mid s)}
    =\underbrace{\log\frac{q_{\mathrm{base}}(v\mid s)}{p_0(v\mid s)}}_{\delta_{\mathrm{pre}}(v\mid s)}
     +\underbrace{\log\frac{q_m(v\mid s)}{q_{\mathrm{base}}(v\mid s)}}_{\delta_m(v\mid s)}.
    \label{eq:teacher_source_decomposition}
\end{equation}
Following log-probability increment extraction~\citep{heo2026opd2},
we apply the post-training changes to student Base to define
\begin{equation}
    q_{\mathrm{inc},m}(v\mid s)
    =\frac{p_0(v\mid s)\exp[\delta_m(v\mid s)]}{Z_m(s)}.
    \label{eq:transferred_teacher}
\end{equation}
Here $Z_m(s)$ normalizes the distribution. Multiplying student Base
probabilities by the post-training ratios $q_m/q_{\mathrm{base}}$
transfers the acquired changes onto the student's own Base, retaining
any SFT response habits. The resulting distribution defines a frozen teacher.

\par\smallskip\noindent\textbf{Post-training the teacher in-domain (A/B and A/D).}
With the student initialization and larger teacher's pretrained starting
point fixed, Table~\ref{tab:teacher_source_abc} compares the Base teacher
(A) and its SFT (B) and RLVR (D) endpoints under the same OPD setup.
In both model pairs, post-training the larger Base teacher with either
SFT or RLVR lowers student $\rho$.
In-domain JS increases
proportionally more than OOD JS, indicating that the acquired preferences
concentrate student changes more strongly in-domain.
Full RLVR teachers yield lower $\rho$ and better OOD retention than
full SFT teachers, supporting a difference associated with how the
teacher acquired its in-domain preferences.

\par\smallskip\noindent\textbf{Transferring post-training changes (B/C and D/E).}
Conditions C and E transfer the SFT and RLVR increments onto student Base,
respectively, to examine whether selective updates require also fitting
the larger Base's predictions.
For SFT (B/C), transferring the post-training increment preserves
nearly the same in-domain JS while reducing OOD JS, $\rho$, and
forgetting; for RLVR (D/E), $\rho$ remains comparably low.
We interpret this contrast through the preferences each teacher acquires
during post-training. The SFT increment includes preferences for the content
and response patterns of external demonstrations~\citep{chu2025sftmemorizes}.
B combines these acquired preferences with imitation of the larger Base's predictions; C's improved
retention indicates that this additional imitation contributes to
the OOD cost of distilling the SFT teacher. Online RLVR improves rewards on its own responses
and tends toward solutions close to its starting
distribution. Its increment
captures these reward-based adjustments to existing behavior.
Our comparison suggests that these task preferences largely determine
the student's relative OOD change, allowing RLVR's conservative
adaptation to carry over through distillation.

\begin{table}[!htbp]
    \centering
    \small
    \setlength{\tabcolsep}{4pt}
    \begingroup
\newcommand{\tablethreerhomark}[2]{\tikz[remember picture,baseline=(#1.base)]{\node[inner sep=0pt,outer sep=0pt] (#1) {#2};}}
\begin{tabular}{@{}lrrrr@{\,}lr@{\,}l@{}}
\toprule
Teacher supervision & \multicolumn{3}{c}{Distribution change} & \multicolumn{4}{c}{Accuracy (\%)} \\
\cmidrule(lr){2-4}\cmidrule(l){5-8}
& $D_{\mathrm{in}}$ & $D_{\mathrm{OOD}}$ & $\rho$ & \multicolumn{2}{c}{Math ($\Delta$)} & \multicolumn{2}{c}{OOD ($\Delta$)} \\
\midrule
\multicolumn{8}{l}{\textbf{Qwen2.5-7B$\to$Qwen2.5-1.5B}} \\
Student Base & & & & 13.93 & \(\phantom{(+00.00)}\) & 35.20 & \(\phantom{(+00.00)}\) \\
A: Larger Base teacher & 0.3696 & 0.1151 & \tablethreerhomark{table-three-rho-0-top}{0.311} & 14.67 & \((+0.75)\) & 34.82 & \((-0.39)\) \\
\addlinespace[3pt]
B: Larger SFT teacher & 4.5711 & 0.9909 & 0.217 & 3.15 & \((-10.78)\) & 34.52 & \((-0.69)\) \\
C: SFT increment & 4.7088 & 0.7879 & 0.167 & 5.20 & \((-8.73)\) & 35.60 & \((+0.39)\) \\
\addlinespace[3pt]
D: Larger RLVR teacher & 4.1650 & 0.6873 & 0.165 & 19.17 & \((+5.25)\) & 35.99 & \((+0.79)\) \\
E: RLVR increment & 4.0247 & 0.5658 & \tablethreerhomark{table-three-rho-0-bottom}{0.141} & 19.22 & \((+5.29)\) & 35.76 & \((+0.56)\) \\
\midrule
\multicolumn{8}{l}{\textbf{Qwen3-8B$\to$Qwen3-4B}} \\
Student Base & & & & 26.85 & \(\phantom{(+00.00)}\) & 45.45 & \(\phantom{(+00.00)}\) \\
A: Larger Base teacher & 0.3556 & 0.2286 & \tablethreerhomark{table-three-rho-1-top}{0.643} & 27.85 & \((+1.01)\) & 45.76 & \((+0.31)\) \\
\addlinespace[3pt]
B: Larger SFT teacher & 5.3947 & 1.1765 & 0.218 & 28.72 & \((+1.87)\) & 43.54 & \((-1.92)\) \\
C: SFT increment & 5.3346 & 0.8362 & 0.157 & 28.84 & \((+1.99)\) & 44.57 & \((-0.88)\) \\
\addlinespace[3pt]
D: Larger RLVR teacher & 3.1320 & 0.4312 & 0.138 & 39.34 & \((+12.49)\) & 45.22 & \((-0.23)\) \\
E: RLVR increment & 2.8076 & 0.3995 & \tablethreerhomark{table-three-rho-1-bottom}{0.142} & 41.54 & \((+14.70)\) & 45.27 & \((-0.19)\) \\
\bottomrule
\end{tabular}%
\begin{tikzpicture}[remember picture,overlay,blend mode=multiply]
\fill[blue!5] ([xshift=-\tabcolsep,yshift=2pt]table-three-rho-0-top.north west) rectangle ([xshift=\tabcolsep,yshift=-2pt]table-three-rho-0-bottom.south east);
\fill[blue!5] ([xshift=-\tabcolsep,yshift=2pt]table-three-rho-1-top.north west) rectangle ([xshift=\tabcolsep,yshift=-2pt]table-three-rho-1-bottom.south east);
\end{tikzpicture}
\endgroup

    \caption{Student distribution changes and capabilities under different
    information sources. All A--E conditions use student step 120.
    JS uses shared prefixes, is relative to Base,
    and is reported in $10^{-2}$ nats. Parentheses show accuracy changes
    from Base in percentage points. Datasets are weighted equally within
    each domain; $\rho$ is undefined at Base.}
    \label{tab:teacher_source_abc}
\end{table}

These comparisons connect the information supplied by teachers to
update selectivity and OOD retention. We next examine how students
inherit learning and retention patterns from SFT and RLVR teachers,
relating these patterns to parameter paths and prediction matching.

\subsection{Parameter and Prediction Trajectories Behind Similar Behavior}
\label{sec:teacher_target}
\label{sec:weight_scaling}

\paragraph{Self-OPD reproduces source-dependent learning and retention patterns.}
We pair each Self-OPD student with the SFT/RLVR run that produced its frozen teacher. Both start from the same Base; the source run provides a reference for the capabilities and behavior the student inherits. Students of RLVR teachers approach
teacher task performance early with similar capability retention.
Students of SFT teachers incur OOD declines approaching or exceeding their teachers'
(Appendix~\ref{app:rawraw_source_curves}).
We examine whether these shared behavioral patterns correspond to
similar parameter paths.

\paragraph{Models along parameter lines provide a capability reference.}
We characterize the learning and retention available along the parameter directions defined by these endpoints.
We first interpolate and extrapolate the SFT, RLVR, and Self-OPD
endpoints~\citep{huang2026peftarena} to examine changes in update magnitude:
\begin{equation}
    \theta_\lambda=\theta_0+\lambda(\theta_{\mathrm{end}}-\theta_0).
    \label{eq:weight_scaling}
\end{equation}
Here $\theta_0$ and $\theta_{\mathrm{end}}$ are the Base and trained
parameters; $\lambda=0$ and $\lambda=1$ recover these two endpoints.
The source SFT/RLVR endpoints are the frozen teachers, so their sweeps
provide a capability reference along the Base--teacher line;
Self-OPD sweeps examine the student's own update direction.

\begin{figure}[!htb]
    \centering
    \includegraphics[width=\textwidth]{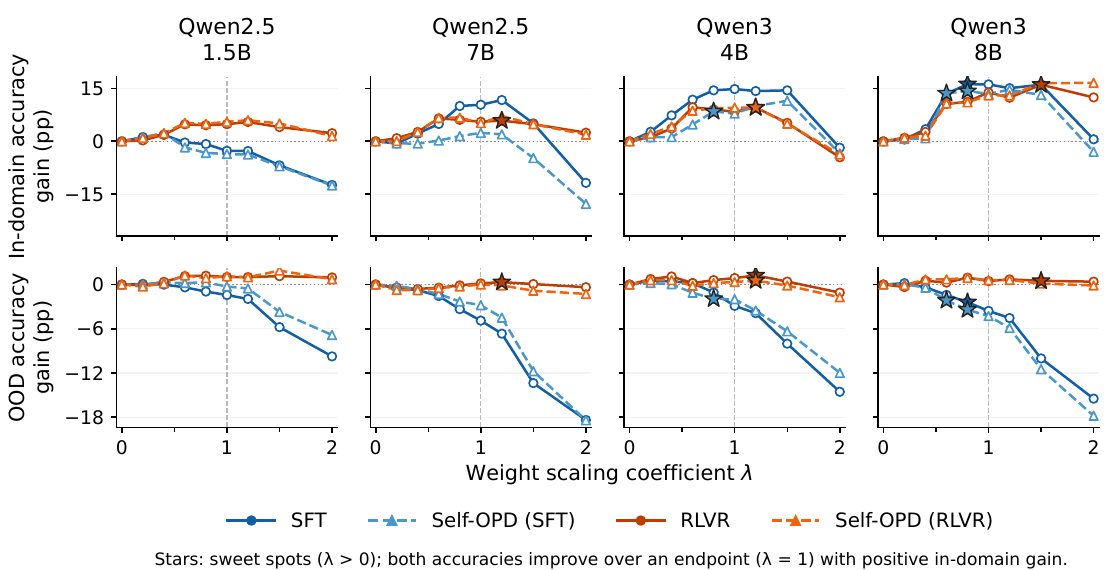}
    \caption{Accuracy changes from Base under weight scaling.
    Stars mark sweet-spot coefficients that improve both accuracies over
    the original endpoint, as defined in
    Appendix~\ref{app:rawraw_weight_scaling}.}
    \label{fig:weight_scaling_accuracy}
\end{figure}

Figure~\ref{fig:weight_scaling_accuracy} shows broader ranges of retained
task gains for RLVR-based updates, with a more consistent advantage in
OOD retention: accuracy remains near Base even for $\lambda>1$, while
forgetting grows faster along SFT-based directions.
Sweet spots occur in interpolation for SFT-based updates and
extrapolation for RLVR-based updates.
These patterns motivate a geometric explanation for Self-OPD: students may inherit teacher behavior by moving along the same Base--teacher parameter direction.

\begin{figure}[!htb]
    \centering
    \includegraphics[width=\textwidth]{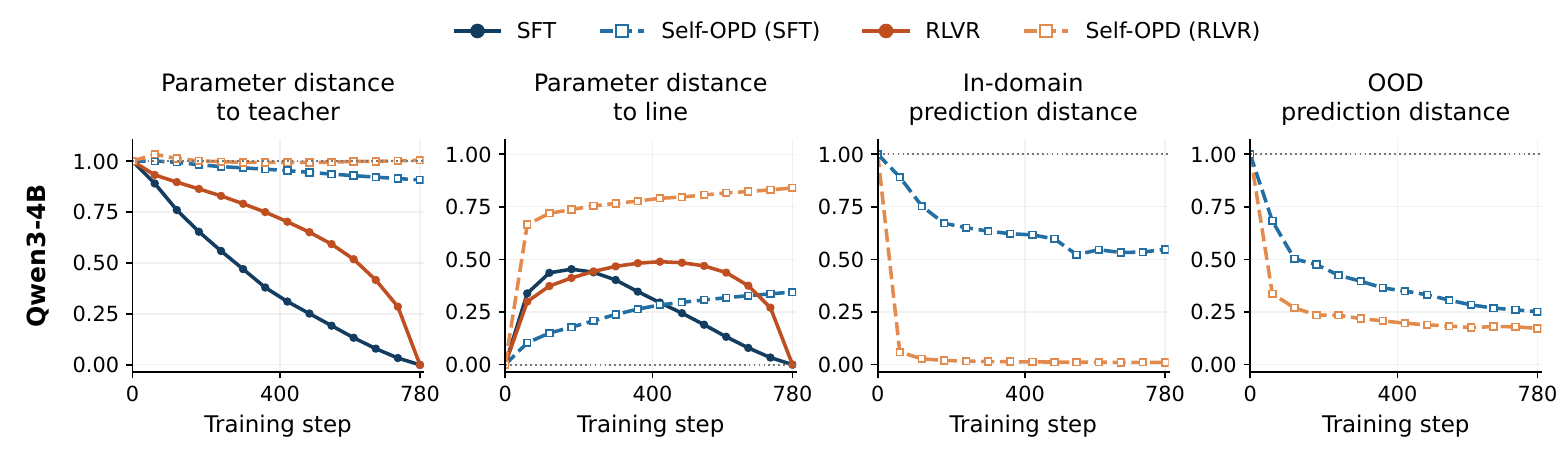}
    \caption{Qwen3-4B parameter distances for SFT, RLVR, and their Self-OPD
    students (left), and student--teacher JS for the same students (right).
    Each distance is normalized by its corresponding initial Base--teacher gap.}
    \label{fig:self_opd_trajectory_comparison}
\end{figure}

\paragraph{Parameter and prediction distances evolve differently.}
We next test whether Self-OPD students actually move along the Base--teacher line.
We measure parameter distances to the teacher, $d_\theta$, and to the
Base--teacher line, $d_\perp$, normalized by the initial parameter gap
(Appendix~\ref{app:teacher_trajectories}). Prediction distance to the
frozen teacher $q_T$ is likewise normalized by its initial value:
\begin{equation}
    d_{\mathrm{out},d}(t)
    =\frac{D_d(p_t,q_T)}{D_d(p_0,q_T)},
    \qquad d\in\{\mathrm{in},\mathrm{OOD}\}.
    \label{eq:teacher_output_distance}
\end{equation}
Figure~\ref{fig:self_opd_trajectory_comparison} (left) shows that Self-OPD
students develop substantial deviations from the Base--teacher line and
remain distant from the teacher's parameters. To examine how this
parameter-space departure relates to behavior, we compare their
predictions with those of the same fixed teachers.
Figure~\ref{fig:self_opd_trajectory_comparison} (right) shows that the
RLVR-teacher student closely approaches teacher predictions in both domains.
Both teacher sources reduce OOD prediction gaps, while SFT-teacher students
generally retain larger in-domain gaps. Together, these measurements show
that students can approach teacher behavior along different parameter
paths, with different degrees of task learning and OOD change inherited
from each source.

Appendix Table~\ref{tab:self_opd_endpoint_distances} reports the corresponding
endpoint distances for the other four models.

\paragraph{Student probabilities shape preference transfer.}
Teacher feedback on prefixes the student rarely generates has fewer
opportunities to contribute to training. This motivates examining how
student probabilities affect the transfer of teacher preferences.
Let $d_{p_t}$ be the training-prefix distribution and
$A_t(v\mid s)=\log q_T(v\mid s)-\log p_t(v\mid s)$ the feedback.
For a full-vocabulary score-function model with fixed sampling
distributions and stop-gradient feedback, the mean gradient is
\begin{equation}
    \begin{aligned}
    \mu_t(v\mid s)
       &=p_t(v\mid s)\bigl(\overline A_t(s)-A_t(v\mid s)\bigr),\\
    \mathbb E[\nabla_\theta\mathcal L]
       &=\mathbb E_{s\sim d_{p_t}}\!\left[
           \sum_v\mu_t(v\mid s)\nabla_\theta z_t(v\mid s)\right],
    \end{aligned}
    \label{eq:conditional_token_gradient}
\end{equation}
where $\overline A_t(s)=\mathbb E_{a\sim p_t}[A_t(a\mid s)]$ and
$z_t$ denotes student logits.
Student prefix and token probabilities weight teacher feedback;
these weights and the logit Jacobians change along the student's path
(Appendix~\ref{app:rawraw_target_theory}).

We first examine how readily students can generate their teachers'
response prefixes. On the same mathematical problems, SFT-teacher prefixes
have lower probability under the shared Base than RLVR-teacher prefixes,
and this difference persists during student training
(Appendix~\ref{app:teacher_prefix_probability}). Since OPD receives feedback
on student-generated prefixes, teacher preferences associated with rarely
visited prefixes have fewer opportunities to contribute to updates.
This offers an explanation for slower transfer of SFT task gains relative
to OOD changes.

Response openings provide a concrete instance of this mismatch. In our
diagnostics, SFT teachers inherit openings such as \texttt{Okay} and
\texttt{Alright} from reference answers, while their preferred opening
tokens fall outside the Base student's Top-16 candidates.
In Self-OPD (SFT), zeroing feedback at the first
response position during training improves final mathematical accuracy by
3.36 percentage points on average across five models relative to ordinary
feedback, while the OOD accuracy difference is at most 0.53 percentage
points per model (Appendix~\ref{app:sft_opening_feedback}). This intervention shows
that feedback tied to teacher response habits can impede task learning.
\FloatBarrier

\section{Conclusion}

OPDArena provides a shared framework for comparing task learning,
capability retention, and predictive distribution changes across SFT,
RLVR, and OPD. Across these methods, OOD distribution change is more
consistently associated with forgetting than in-domain change,
motivating $\rho$ as a measure of update selectivity.
Controlled comparisons of OPD teachers show that the preferences they
acquire through in-domain post-training lead to more selective student
updates.
The student's own distribution further shapes how these preferences
are inherited. These findings motivate selecting OPD teachers by the
information they add relative to student Base and its interaction with
student predictions, alongside teacher capabilities.

\bibliographystyle{plainnat}
\bibliography{references}

\appendix
\makeatletter
\setlength{\@fptop}{0pt}
\setlength{\@fpsep}{12pt plus 2pt minus 2pt}
\setlength{\@fpbot}{0pt plus 1fil}
\setlength{\@dblfptop}{0pt}
\setlength{\@dblfpsep}{12pt plus 2pt minus 2pt}
\setlength{\@dblfpbot}{0pt plus 1fil}
\makeatother
\clearpage
\section{Related Work}
\label{sec:related}

\paragraph{Learning and forgetting in post-training.}
Acquiring new skills while retaining prior capabilities is a central
problem in continual learning~\citep{wang2024continualsurvey}.
Empirical studies show that language-model fine-tuning can impair
capabilities beyond the target task, motivating evaluations that
jointly measure adaptation and retention~\citep{luo2025forgetting}.
For language-model post-training, prior work connects the retention
advantage of online RL to limited KL changes on the new
task~\citep{shenfeld2025rlrazor}. A complementary study of multimodal
language models analyzes how supervision distributions affect learning
and interference with prior knowledge~\citep{zhang2026rft}.
Our comparison examines this learning--forgetting relationship under
token-level supervision from different teachers.

\paragraph{On-policy distillation.}
Training on student-generated trajectories exposes a distillation
teacher to the contexts its student actually visits, reducing the
mismatch between training and deployment
distributions~\citep{agarwal2024onpolicy}.
MiniLLM uses reverse-KL distillation to align student and teacher
predictions~\citep{gu2024minillm}. Subsequent analysis identifies
compatible thinking patterns and additional teacher capabilities as
important conditions for successful OPD, and traces learning to
alignment on high-probability tokens at student-visited
states~\citep{li2026rethinkingopd}.
Related work also finds that smaller models can struggle to learn
from strong reasoning demonstrations~\citep{li2025smallmodels}.
These studies motivate our analysis of how teacher preferences affect
student predictions across domains.

\paragraph{Self-distillation and teacher information.}
Teacher information can enter through both model parameters and
conditioning context. In OPSD, access to a reference solution provides the teacher with
privileged information hidden from the student~\citep{zhao2026opsd}.
On-policy context distillation similarly transfers information
available in a teacher's context into student
parameters~\citep{ye2026opcd}.
Teacher conditioning, update rules, and the response tokens receiving
supervision influence the resulting training
signal~\citep{zhao2026opsd,ye2026opcd,wang2026denser}.
On-policy delta distillation extracts post-training information through
teacher log-probability differences~\citep{heo2026opd2}.
We use teacher comparisons and this decomposition to examine how contextual,
pretrained, and post-trained preferences affect update selectivity.

\paragraph{Distribution changes and parameter paths.}
Prediction and parameter measurements provide complementary views of
post-training. Distribution-based analyses connect supervision to
changes in model behavior~\citep{zhang2026rft}.
Complementary work jointly evaluates learning and retention across
fine-tuning parameterizations and uses interpolation to study behavior
along adaptation paths~\citep{huang2026peftarena}.
We combine parameter and prediction measurements to examine how Self-OPD
students acquire teacher behavior along their own training paths.

\clearpage
\clearpage
\section{Experimental Details}
\label{app:experimental_details}

\subsection{Base-Model Baselines}
\label{app:base_model_baselines}

Tables~\ref{tab:base_indomain} and~\ref{tab:base_ood} report standalone
in-domain and OOD evaluations of the released Base models. Table~\ref{tab:base_indomain}
compares Raw inputs with the corresponding Qwen chat template;
Table~\ref{tab:base_indomain_lengths} reports the corresponding output lengths.
Qwen's documentation describes Base releases as pretrained models that
have not learned the predefined chat template.\footnote{Qwen documentation,
``Key Concepts'': \url{https://qwen.readthedocs.io/en/latest/getting_started/concepts.html}.}
Raw input yields higher mean in-domain accuracy than Template for every
model in the paired comparison, supporting the use of the Base models'
text-completion format. We use Raw input for the models being trained
and their evaluations.
Accuracy values are percentages.
For in-domain and OOD, training curves use a shared original-model baseline for
each model and input configuration at the start of training.

Our target in-domain evaluation protocol specifies eight responses per
problem at temperature 0.7 and top-$p$ 0.95 for both the original models
and training checkpoints.
The input prompt is limited to 1,024 tokens. For AMC23, AIME24, AIME25, and
HMMT25, each response is limited to 31,744 tokens, within a total context budget
of 32,768 tokens. Math500 uses all 500 test problems with a response limit of
16,384 tokens and a total context budget of 17,408 tokens. Its evaluations use
the Base-model Raw and Template configurations in Table~\ref{tab:base_indomain}, as well as
every available saved checkpoint of the five Base models' Raw training runs.
Under this protocol, acc@8 is the fraction of correct responses averaged
over eight samples and all problems in each dataset. Dataset sizes and
scoring rules are listed in
Appendix~\ref{app:validation_data}.

\begin{table*}[!htb]
\centering
\caption{In-domain acc@8 (\%) of the released Base models.
The Mean column weights the five datasets equally. Raw and Template denote
inputs without and with the corresponding Qwen chat template.}
\label{tab:base_indomain}
\small
\setlength{\tabcolsep}{3pt}
\resizebox{\textwidth}{!}{%
\begin{tabular}{llrrrrrr}
\toprule
Model & Input & AMC23 & AIME24 & AIME25 & HMMT25 & Math500 & Mean \\
\midrule
Qwen2.5-1.5B Base & Raw & 20.00 & 1.25 & 0.83 & 0.42 & 47.13 & 13.93 \\
Qwen2.5-1.5B Base & Template & 1.88 & 0.42 & 0.00 & 0.00 & 4.58 & 1.37 \\
Qwen2.5-7B Base & Raw & 46.88 & 9.17 & 5.00 & 2.08 & 67.28 & 26.08 \\
Qwen2.5-7B Base & Template & 31.25 & 6.67 & 3.33 & 0.42 & 57.03 & 19.74 \\
Qwen2.5-14B Base & Raw & 51.25 & 10.42 & 7.08 & 2.50 & 73.23 & 28.90 \\
Qwen3-1.7B Base & Raw & 32.50 & 4.17 & 4.58 & 0.00 & 59.58 & 20.17 \\
Qwen3-1.7B Base & Template & 14.38 & 2.08 & 0.42 & 0.42 & 25.75 & 8.61 \\
Qwen3-4B Base & Raw & 45.00 & 9.58 & 6.25 & 1.67 & 71.73 & 26.85 \\
Qwen3-4B Base & Template & 30.00 & 8.75 & 4.17 & 1.67 & 37.13 & 16.34 \\
Qwen3-8B Base & Raw & 57.50 & 12.08 & 11.25 & 0.00 & 75.78 & 31.32 \\
Qwen3-8B Base & Template & 48.75 & 12.08 & 10.00 & 2.50 & 67.50 & 28.17 \\
Qwen3-14B Base & Raw & 60.00 & 14.58 & 12.08 & 2.50 & 79.43 & 33.72 \\
\bottomrule
\end{tabular}%
}
\end{table*}

\begin{table*}[!ht]
\centering
\caption{Mean in-domain output length in tokens for the same Base-model evaluations.
The Mean column weights the five datasets equally. Math500 uses a
16,384-token response limit; the other datasets use 31,744 tokens.}
\label{tab:base_indomain_lengths}
\normalsize
\setlength{\tabcolsep}{3pt}
\begin{tabular*}{\textwidth}{@{\extracolsep{\fill}}llrrrrrr}
\toprule
Model & Input & AMC23 & AIME24 & AIME25 & HMMT25 & Math500 & Mean \\
\midrule
Qwen2.5-1.5B Base & Raw & 2359.4 & 5101.0 & 3827.6 & 3450.3 & 1074.3 & 3162.5 \\
Qwen2.5-1.5B Base & Template & 25725.3 & 25467.4 & 25473.3 & 22321.9 & 13228.8 & 22443.4 \\
Qwen2.5-7B Base & Raw & 1160.0 & 3131.1 & 3121.3 & 2364.9 & 787.3 & 2112.9 \\
Qwen2.5-7B Base & Template & 1541.7 & 1985.4 & 2171.4 & 2038.7 & 696.8 & 1686.8 \\
Qwen2.5-14B Base & Raw & 912.9 & 1214.1 & 1483.7 & 1317.1 & 614.7 & 1108.5 \\
Qwen3-1.7B Base & Raw & 1807.0 & 5101.9 & 4117.2 & 4692.0 & 1073.2 & 3358.3 \\
Qwen3-1.7B Base & Template & 11554.2 & 14284.5 & 14597.9 & 12014.1 & 5884.6 & 11667.1 \\
Qwen3-4B Base & Raw & 1399.7 & 2901.5 & 1803.1 & 1610.1 & 719.4 & 1686.8 \\
Qwen3-4B Base & Template & 11628.7 & 11135.5 & 11435.0 & 10028.8 & 6385.6 & 10122.7 \\
Qwen3-8B Base & Raw & 1082.4 & 2600.4 & 2191.7 & 1612.8 & 711.8 & 1639.8 \\
Qwen3-8B Base & Template & 2077.9 & 4310.8 & 4274.0 & 3950.0 & 1068.1 & 3136.2 \\
Qwen3-14B Base & Raw & 1205.0 & 2291.0 & 1386.7 & 1060.6 & 663.0 & 1321.2 \\
\bottomrule
\end{tabular*}
\end{table*}

\begin{table*}[!ht]
\centering
\caption{OOD accuracy (\%) of the released Base models with Raw input. The macro is an unweighted mean over the
seven datasets.  HellaSwag, ARC-Challenge, and MMLU use candidate likelihood;
the other tasks score one generated response. Response budgets are listed
in Table~\ref{tab:validation_suite}.}
\label{tab:base_ood}
\small
\setlength{\tabcolsep}{4pt}
\resizebox{\textwidth}{!}{%
\begin{tabular}{lrrrrrrrr}
\toprule
Model & HellaSwag & ARC-C & MMLU & IFEval & TriviaQA & BBH & NQ-Open & Macro \\
\midrule
Qwen2.5-1.5B Base & 49.22 & 39.45 & 61.40 & 20.70 & 22.00 & 42.59 & 11.05 & 35.20 \\
Qwen2.5-7B Base & 61.33 & 44.53 & 74.21 & 31.79 & 44.40 & 72.96 & 19.53 & 49.82 \\
Qwen2.5-14B Base & 62.11 & 50.78 & 78.07 & 31.79 & 55.20 & 79.63 & -- & -- \\
Qwen3-1.7B Base & 46.09 & 37.11 & 65.26 & 24.03 & 20.60 & 50.37 & 9.17 & 36.09 \\
Qwen3-4B Base & 53.13 & 41.02 & 75.09 & 31.98 & 34.00 & 66.67 & 16.29 & 45.45 \\
Qwen3-8B Base & 59.77 & 44.53 & 76.49 & 37.52 & 47.20 & 72.59 & 21.50 & 51.37 \\
Qwen3-14B Base & 61.72 & 48.44 & 80.70 & 43.25 & 54.80 & 79.26 & -- & -- \\
\bottomrule
\end{tabular}%
}
\end{table*}

\subsection{Training Input Formats}
\label{app:training_input_formats}

All methods train on OpenR1-50K.

\paragraph{Rollout input format.}
All five Base models use Raw input for their rollout-based methods.
The model receives only the original problem text:
\begin{verbatim}
{ORIGINAL_PROBLEM}
\end{verbatim}
No system message, user/assistant delimiters, chat generation prefix, or
empty thinking block is inserted.  Student rollouts and in-domain validation
use the same Raw protocol.
For Self-OPD and OPD from larger models, teachers also use Raw input
when scoring student-generated responses.

\paragraph{SFT input format.}
Raw SFT tokenizes the original problem as the prompt and its complete worked
solution as the target, without chat-control tokens.  An end-of-sequence token
is appended to the solution.  Cross-entropy loss is applied only to solution
and end-of-sequence tokens; prompt tokens are masked.  This convention applies
to both Qwen2.5 and Qwen3 Base students.

\paragraph{OPSD teacher augmentation.}
The teacher receives the original problem with the following private block;
the student receives only the original user content:
\begingroup\footnotesize
\begin{verbatim}
Here is a privileged reference solution to the original problem:
=== Reference Solution Begin ===

{FULL_REFERENCE_SOLUTION}

=== Reference Solution End ===

Use the reference solution as private guidance. Independently solve the
original problem, and produce the answer in the format requested in the
original prompt.
\end{verbatim}
\endgroup

For the OPSD accuracy comparisons, we append
\texttt{Answer to the original problem:} on a new line after the private
block and before the student response. This preserves the explicit
transition supplied by the original OPSD teacher's chat prompt while
using the Raw text-completion format of our Base models~\citep{zhao2026opsd}.
The student receives only the Raw problem, and the teacher scores its tokens.
Appendix~\ref{app:opsd_formats} compares three teacher input formats,
their answer-boundary diagnostics, and full training results, and
specifies the Raw-teacher configuration used for the distribution comparison.

\subsection{Evaluation Input Formats}
\label{app:evaluation_input_formats}

\paragraph{In-domain.}
Raw tokenizes the dataset user content directly without chat control tokens.
For example, the complete Raw input for the first AIME25 validation item is:
\begingroup\footnotesize
\begin{verbatim}
Find the sum of all integer bases $b>9$ for which $17_{b}$ is a divisor
of $97_{b}$. Please reason step by step, and put your final answer
within \boxed{}.
\end{verbatim}
\endgroup
The in-domain evaluations along the training curves use this Raw format.
Standalone baseline input modes are identified in the baseline tables.

\paragraph{OOD.}
\label{app:ood_prompts}

All seven OOD tasks use task-native Raw input without a chat template.

HellaSwag scores candidate endings appended to the activity and context.
ARC-Challenge scores each answer after the question and choices. MMLU uses
zero-shot subject prompts and scores answer letters. IFEval checks the official
instruction constraints; TriviaQA uses flexible answer containment. BBH uses
three chain-of-thought examples and exact-matches the extracted final answer,
stopping at EOS or a newline followed by \texttt{Q:}; paragraph breaks are
allowed. We extract the last explicit conclusion, accepting capitalization,
colon, ``correct/final answer,'' and boxed-answer variants. For multiple-choice
tasks we retain the option label; for other tasks we extract the short answer
in the required format, removing quotes and accompanying explanations. Otherwise,
we use the text extracted after the first literal ``the answer is'' cue.

\subsection{Training Settings}
\label{app:training_settings}

Table~\ref{tab:training_settings} summarizes the training configurations.
We use separate learning rates for SFT and on-policy methods, following
prior training setups~\citep{li2026rethinkingopd,huang2026peftarena}.
We train all methods on OpenR1-50K, a subset of the Open R1
data~\citep{huggingface2025openr1}, and evaluate on five in-domain and seven
OOD tasks. The target in-domain evaluation protocol specifies eight
responses per problem across all five datasets.

Our student Top-16 setting follows the ablation of
\citet{li2026rethinkingopd}: increasing $k$ beyond $4$ brings negligible
additional accuracy gains, while Top-16 and Top-64 maintain smooth
training. Top-16 retains this stability with lower candidate-processing
overhead than Top-64.

\begin{table*}[t]
\centering
\caption{Current OpenR1-50K training settings.  OPD, Self-OPD, and OPSD share the listed
optimization settings; the teacher input or token signal differs.}
\label{tab:training_settings}
\small
\setlength{\tabcolsep}{4pt}
\resizebox{\textwidth}{!}{%
\begin{tabular}{llll}
\toprule
Setting & SFT & RLVR & OPD variants / OPSD \\
\midrule
Train records & 49,993 & 49,993 & 49,993 \\
Epochs & 1 & 1 & 1 \\
Global batch & 64 & 64 & 64 \\
Learning rate & $10^{-5}$ & $10^{-6}$ & $10^{-6}$ \\
Optimizer & AdamW & AdamW & AdamW \\
Adam $(\beta_1,\beta_2)$ & $(0.9,0.999)$ & $(0.9,0.999)$ & $(0.9,0.999)$ \\
Adam $\epsilon$ & $10^{-8}$ & $10^{-8}$ & $10^{-8}$ \\
Weight decay & 0 & 0.01 & 0.01 \\
Global gradient-norm clip & 1.0 & 1.0 & 1.0 \\
Schedule & cosine; 5\% warmup & constant; no warmup & constant; no warmup \\
Training precision & BF16 & FP32 FSDP model dtype & FP32 FSDP model dtype \\
Sequence limits & 14,336 total & 1,024 prompt; 7,168 response & 1,024 prompt; 7,168 response \\
Train responses per prompt & -- & 8 & 4 student responses \\
Train sampling & -- & temp. 1.0; top-$p$ 1.0 & student temp. 1.0; top-$p$ 1.0 \\
Objective & assistant-token CE & token-mean GRPO; KL 0 & token-mean Top-16 teacher signal; KL 0 \\
Teacher support & -- & -- & student top-16; student-$p$ weighting \\
Validation / save interval & 20 / 20 steps & 20 / 20 steps & 20 / 20 steps \\
GPUs per node & 8 A100 GPUs & 8 A100 GPUs & 8 A100 GPUs \\
\bottomrule
\end{tabular}%
}
\end{table*}

For RLVR, correctness is computed by the mathematical rule scorer and GRPO
advantages are formed within each eight-response group.
For OPD and OPSD, we evaluate the teacher on student-generated prefixes
and use the student's Top-16 candidate tokens.
Student probabilities normalized on that support weight the
teacher--student log-probability difference; both log probabilities
are evaluated under the full-vocabulary distributions.
The prepared weights and coefficients are held fixed during the
actor's probability-ratio update.
Self-OPD uses the final saved checkpoints of the corresponding SFT
and RLVR training; teacher parameters remain frozen throughout distillation.

The main results in Tables~\ref{tab:main_results}--\ref{tab:teacher_capability_counterexample}
and Figures~\ref{fig:overview}(b)--\ref{fig:predictive_distributions}
are averaged over two training seeds, as are the supplementary curves in
Appendix~\ref{app:complete_raw} and
Figures~\ref{fig:source_training_accuracy} and~\ref{fig:distribution_forgetting_q3_1p7b}.
For Self-OPD (SFT) and Self-OPD (RLVR), both training stages are repeated:
SFT/RLVR teachers are trained with different seeds, and each frozen endpoint
is used to train a new student from Base, with rollout seeds also varied
between repeats.

\subsection{Validation Data and Scoring}
\label{app:validation_data}

AMC23 and AIME24/AIME25 use problems from the MAA mathematics
competitions,\footnote{\url{https://maa.org/math-competitions/}}
and HMMT25 uses the February HMMT
competition.\footnote{\url{https://www.hmmt.org/www/archive/}}
Math500 is the evaluation subset of MATH~\citep{hendrycks2021math}
released by \citet{lightman2024verify}.
The OOD suite comprises HellaSwag~\citep{zellers2019hellaswag},
ARC-Challenge~\citep{clark2018arc}, MMLU~\citep{hendrycks2021mmlu},
IFEval~\citep{zhou2023ifeval}, TriviaQA~\citep{joshi2017triviaqa},
BBH~\citep{suzgun2023bbh}, and the open-domain split of Natural
Questions~\citep{kwiatkowski2019natural,lee2019latent}.
We compute each group mean only when all five in-domain or all seven OOD
dataset accuracies are available, weighting datasets equally.

\begin{table*}[t]
\centering
\caption{Target evaluation protocol: fixed validation suite and response
budgets in tokens. OOD category
subsamples use seed 42 and are reused at every checkpoint. A dash denotes
likelihood scoring without response generation. The IFEval budget applies
to the response prefix used for scoring.}
\label{tab:validation_suite}
\small
\resizebox{\textwidth}{!}{%
\begin{tabular}{lllrrl}
\toprule
Domain & Dataset & Evaluation & Examples & \shortstack{Response budget\\(tokens)} & Primary score \\
\midrule
\multirow{5}{*}{In-domain}
 & AMC23 & eight generated responses & 40 & 31,744 & acc@8 \\
 & AIME24 & eight generated responses & 30 & 31,744 & acc@8 \\
 & AIME25 & eight generated responses & 30 & 31,744 & acc@8 \\
 & HMMT25 & eight generated responses & 30 & 31,744 & acc@8 \\
 & Math500 & eight generated responses & 500 & 16,384 & acc@8 \\
\midrule
\multirow{7}{*}{OOD}
 & HellaSwag & candidate likelihood & 256 & -- & normalized accuracy \\
 & ARC-Challenge & candidate likelihood & 256 & -- & normalized accuracy \\
 & MMLU & candidate likelihood; 10 per subject & 570 & -- & accuracy \\
 & IFEval & generation & 541 & 512 & prompt-strict accuracy \\
 & TriviaQA & generation & 500 & 256 & flexible answer containment \\
 & BBH & three-shot chain-of-thought; 10 per task & 270 & 1,024 & final-answer exact match \\
 & NQ-Open & five-shot generation & 3,610 & 256 & normalized exact match \\
\bottomrule
\end{tabular}
}
\end{table*}

For IFEval, we retain the first 512 tokens of each generated response and
apply the official prompt-level strict checks. This budget is shared by all
methods, checkpoints, and original-model baselines, including the weight
interpolation and extrapolation experiments. Answer lengths are measured on
the retained text. All four generative OOD tasks sample one response per
problem at temperature 0.7 and top-$p$ 0.95, with the budgets in
Table~\ref{tab:validation_suite}. HellaSwag, ARC-Challenge, and MMLU score
the provided candidate answers without generating a response.

IFEval and NQ-Open use their full evaluation splits. The other OOD tasks use
fixed subsets, with per-subject or per-task sampling where specified in
Table~\ref{tab:validation_suite}. NQ-Open uses five fixed question--answer
examples from its training split. Generation stops at a newline, period,
or comma. We report exact match against any
official answer alias after lowercasing, removing punctuation and English
articles, and normalizing whitespace.

Math evaluation prompts end with ``Please reason step by step, and put your
final answer within \texttt{\textbackslash boxed\{\}}.'' OpenR1-50K training
prompts end with ``Let's think step by step and output the final answer within
\texttt{\textbackslash boxed\{\}}.''

\subsection{Predictive-Distribution Measurements}
\label{app:distribution_measurements}

For each Base student, we construct a fixed prefix bank for all twelve
datasets: five in-domain mathematics benchmarks and seven OOD benchmarks.
We select 16 prompts per dataset and continue each with Base and an RLVR
checkpoint for up to 128 tokens. Up to 16 positions per continuation are
selected; the prompt and preceding response tokens at each position form
a text prefix. All compared methods are evaluated on these same prefixes.

At each prefix, we compute JS divergence between full-vocabulary next-token
distributions, using natural logarithms. We average over the fixed prefixes
within each dataset and then weight datasets equally within each task group,
as defined in Section~\ref{sec:distribution_measures}.
Base comparisons measure the change from initialization; comparisons with
the respective SFT or RLVR teacher measure the distance between Self-OPD
students and their teachers. The in-domain and OOD JS means require
measurements for all five and all seven tasks, respectively.

For each model and method, Table~\ref{tab:main_results} and
Figure~\ref{fig:overview} report step-780 results.
The table gives accuracies and changes from Base; the figure pairs
in-domain gains with OOD accuracy gains.
The overall forgetting hierarchy weights models equally, averaging
larger-teacher configurations within each model before averaging across models.
Figure~\ref{fig:predictive_distributions} reports step-780 JS.
Main-run distribution measurements are taken at saved checkpoints at
intervals of 60 steps; correlations pair JS and accuracy at the same checkpoint.

\clearpage
\section{Complete Raw Results}
\label{app:complete_raw}

\begin{figure}[H]
    \centering
    \includegraphics[width=\textwidth,height=.76\textheight,keepaspectratio]{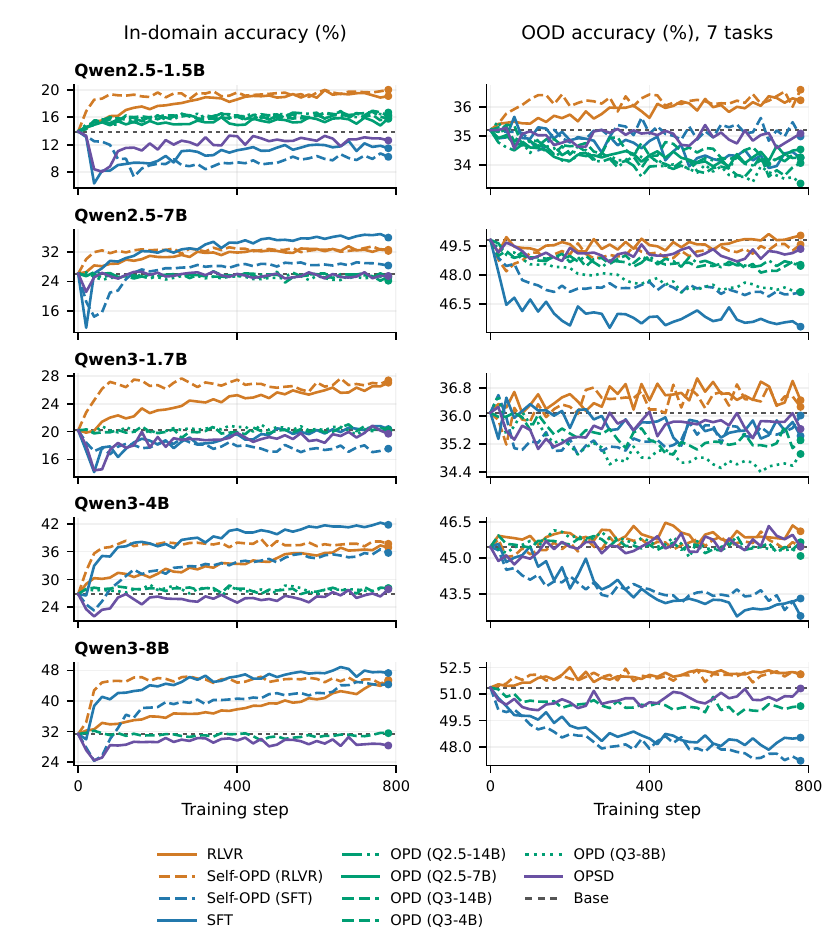}
    \caption{In-domain learning and OOD retention throughout training.
    Curves show task-group mean accuracy; dashed horizontal lines mark
    Base. RLVR and Self-OPD (RLVR) combine in-domain learning with
    sustained capability retention.}
    \label{fig:training_curves}
\end{figure}

Figure~\ref{fig:training_curves} gives the training trajectories across
all five Base models, methods, and teacher configurations, using the
same task groups and Base references as Table~\ref{tab:main_results}.
OPD, Self-OPD, and OPSD use Raw teachers; the OPSD teacher includes the
explicit answer boundary described in Appendix~\ref{app:opsd_formats}.

\clearpage
Each figure below summarizes one Base student, combining in-domain and OOD
evaluations. Panels show the five mathematical and seven OOD tasks;
aggregate accuracies appear in Figure~\ref{fig:training_curves}.
Colors and line styles follow the overall method comparison.
\rawresultfigure{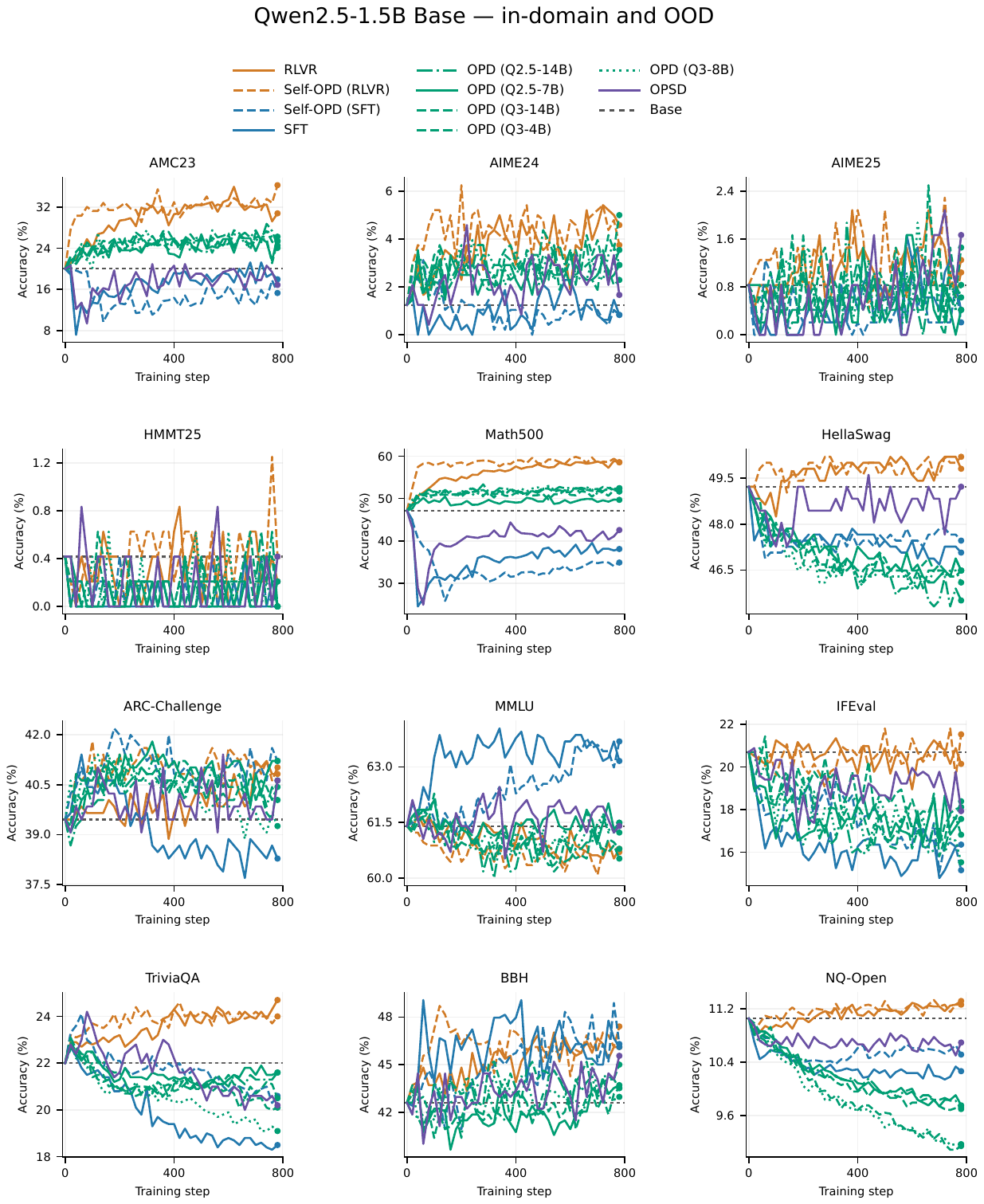}{Qwen2.5-1.5B}{qwen25-15b}
\rawresultpage{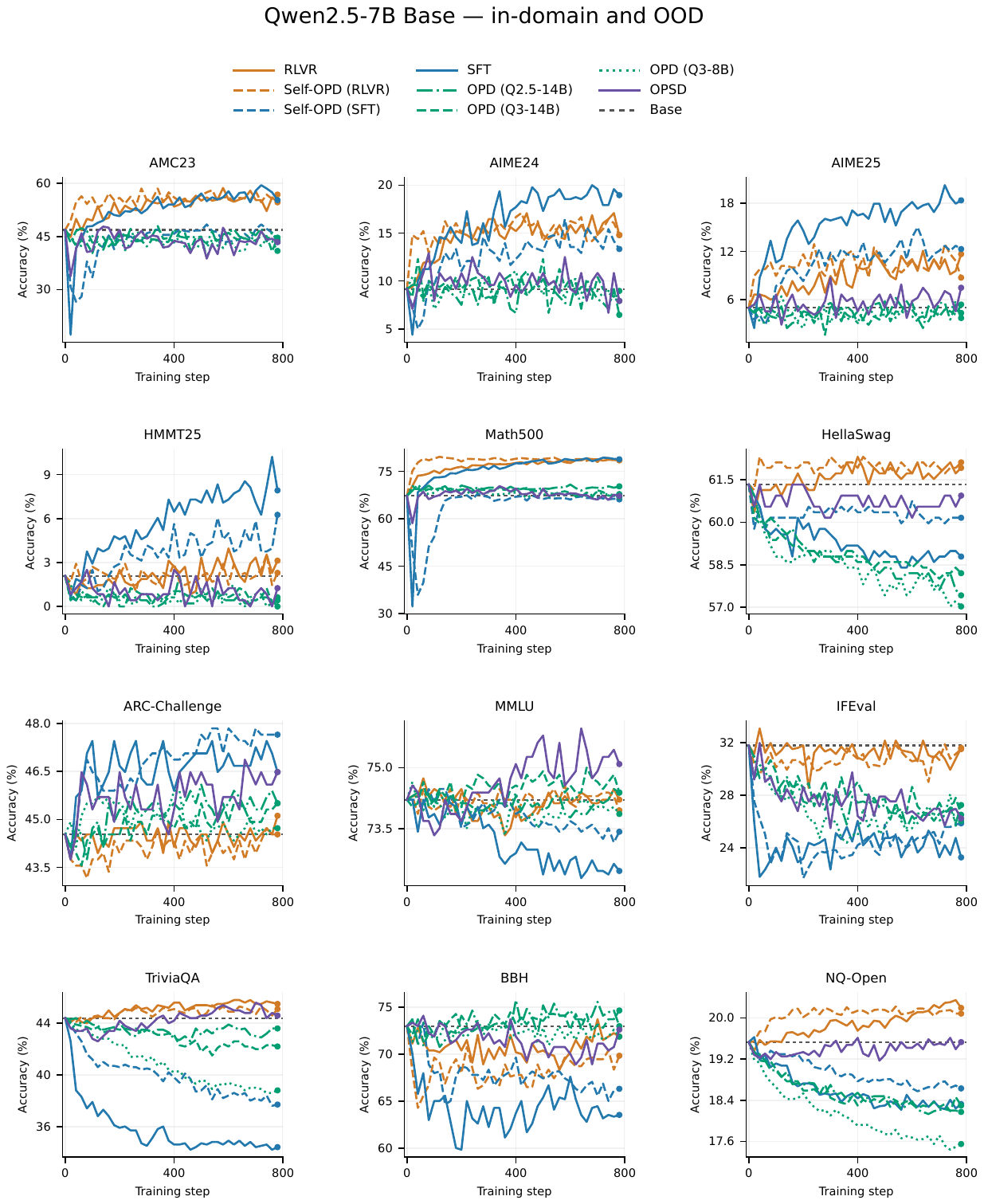}{Qwen2.5-7B}{qwen25-7b}
\rawresultpage{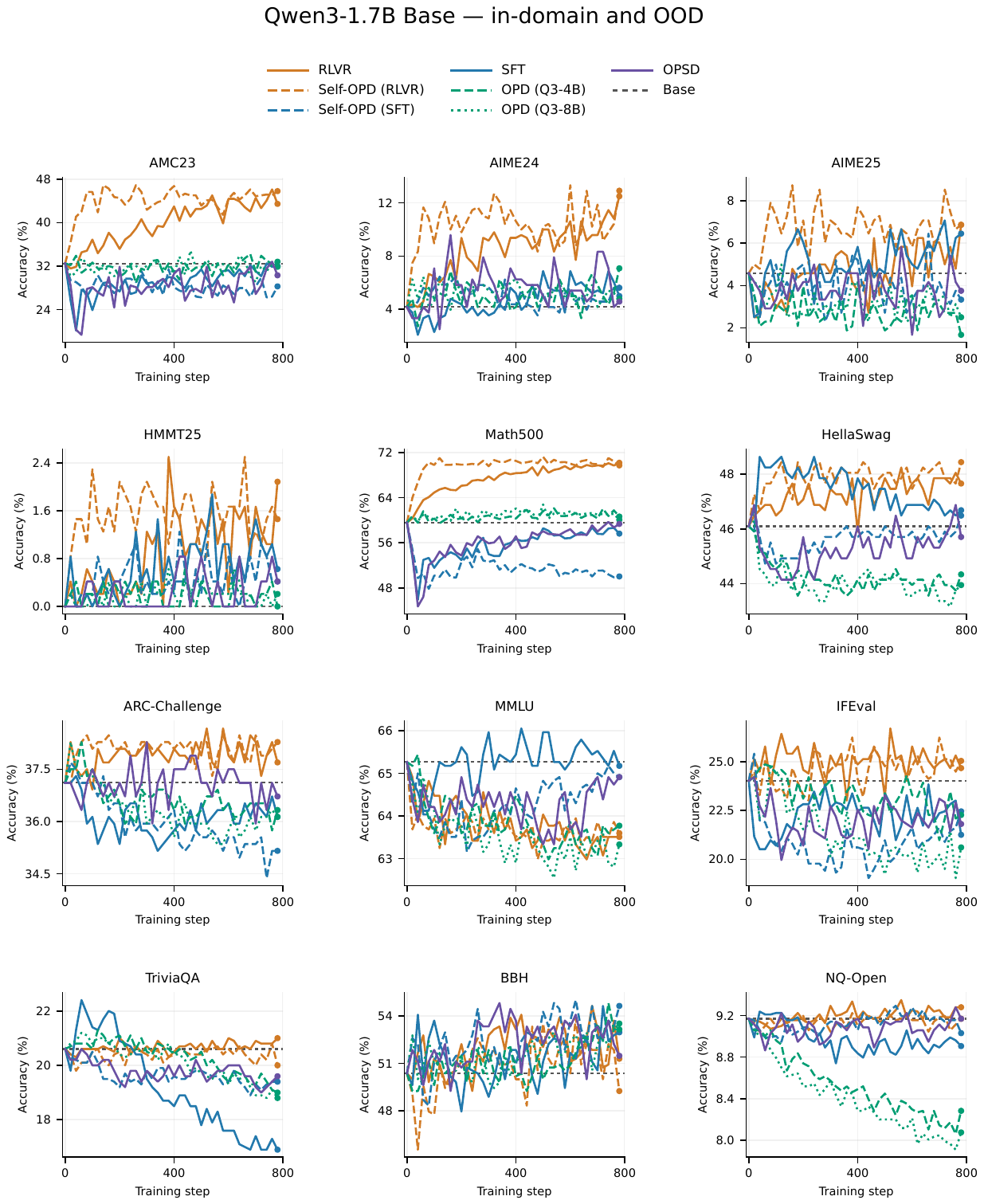}{Qwen3-1.7B}{qwen3-17b}
\rawresultpage{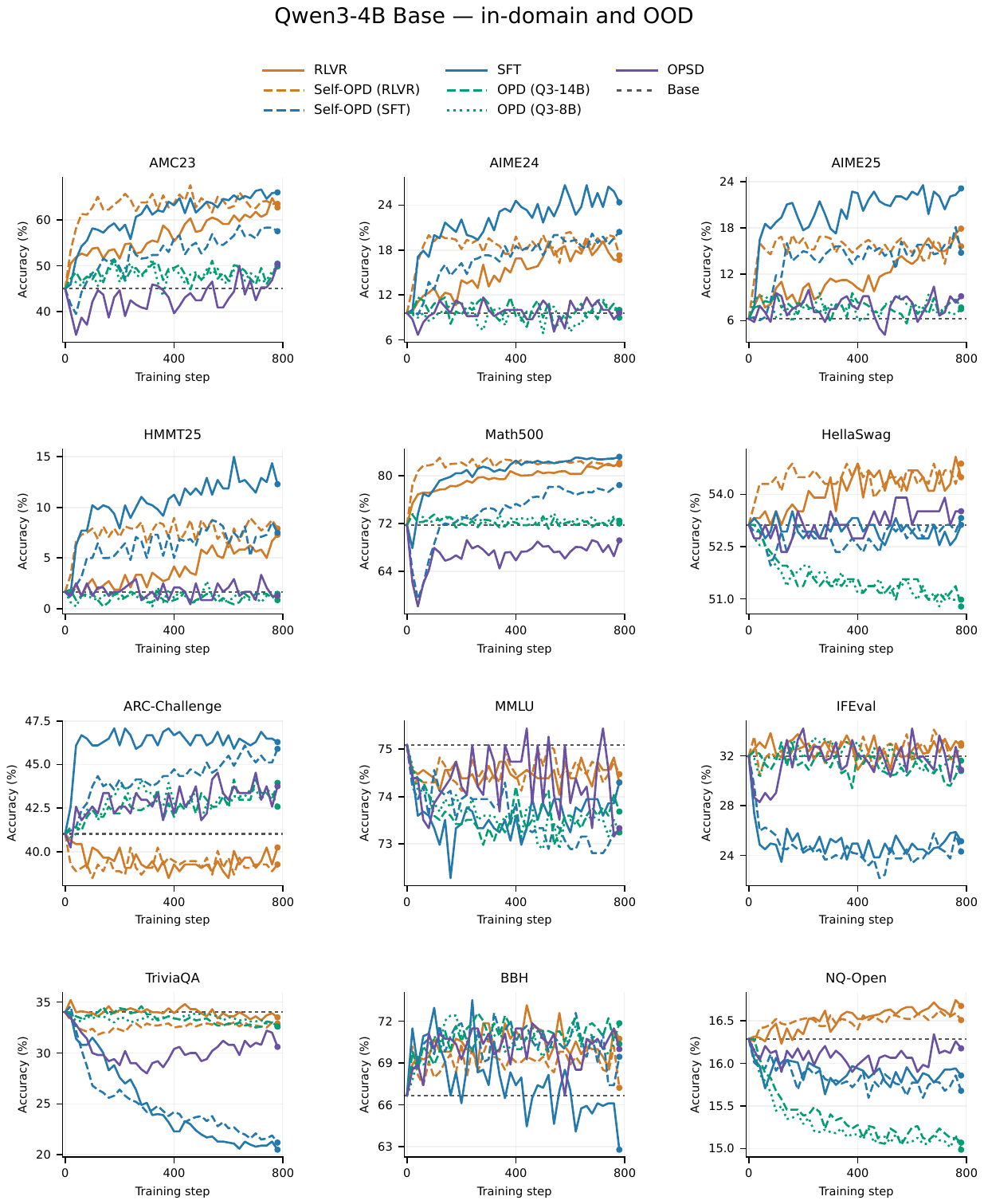}{Qwen3-4B}{qwen3-4b}
\rawresultpage{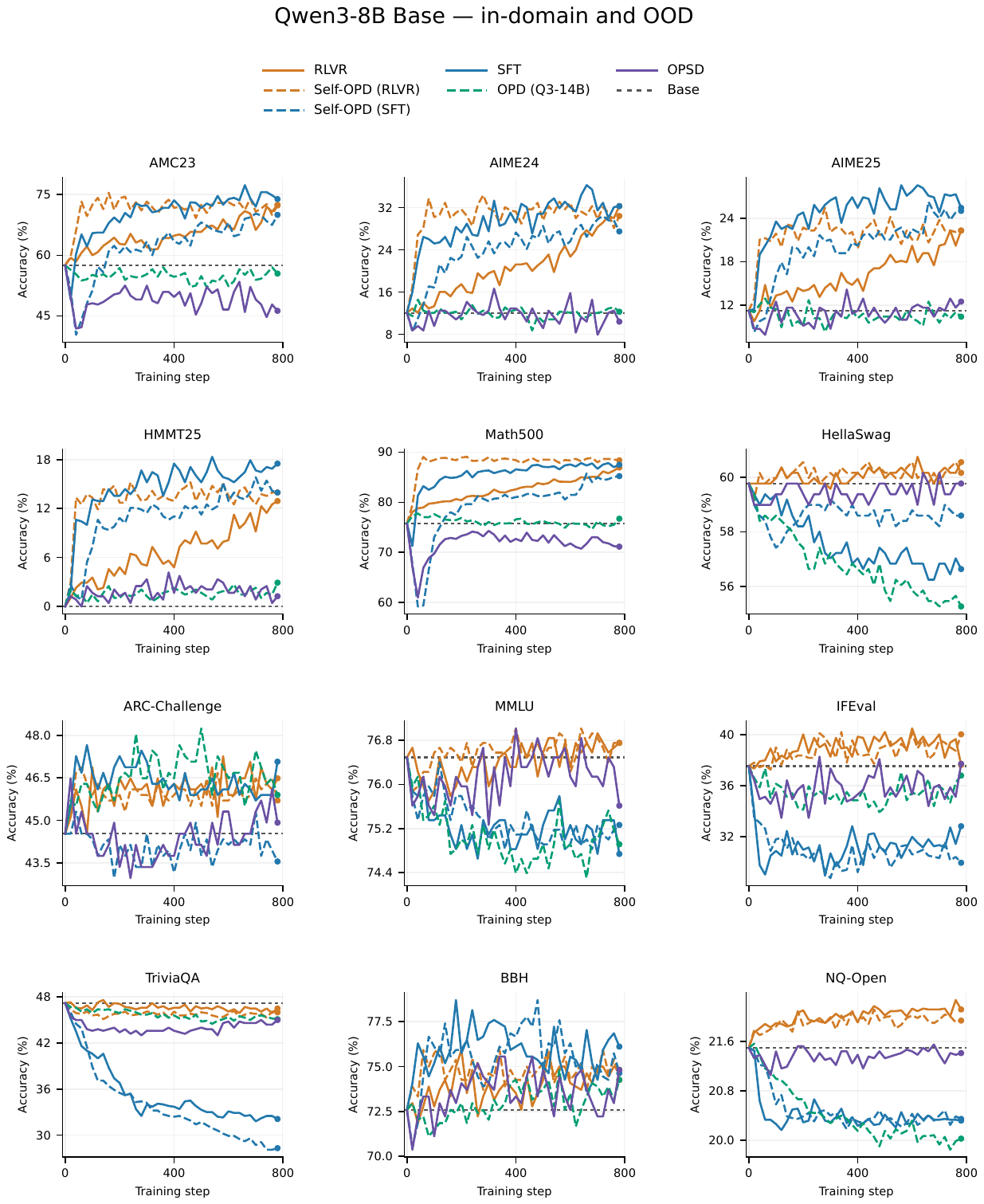}{Qwen3-8B}{qwen3-8b}

\clearpage
\section{Supplementary Measurements for Analysis}
\label{app:supplementary_studies}

\subsection{Student--Teacher Prediction Gaps}
\label{app:rawraw_alignment}
\label{app:student_teacher_alignment}

Let $B,T,S_t$ denote student Base, the frozen teacher, and the current
student at training step $t$. The right panels of
Figure~\ref{fig:self_opd_trajectory_comparison} track the normalized
prediction distance defined in Eq.~\eqref{eq:teacher_output_distance}:
\begin{equation}
    d_{\mathrm{out},d}(t)=\frac{D_d(S_t,T)}{D_d(B,T)},
    \qquad d\in\{\mathrm{in},\mathrm{OOD}\}.
\end{equation}
Lower $d_{\mathrm{out},d}$ means closer teacher predictions relative to
the initial gap. The reduction in that gap is $1-d_{\mathrm{out},d}$;
changes from Base are separately measured by $D_d(B,S_t)$.

On shared evaluation prefixes, the square root of JS satisfies the
triangle inequality, giving
\begin{equation}
    D_{\mathrm{in}}(B,S_t)\geq
    \left[\sqrt{D_{\mathrm{in}}(B,T)}-
    \sqrt{D_{\mathrm{in}}(S_t,T)}\right]_+^2.
\end{equation}
A teacher far from Base and a student close to that teacher therefore
imply substantial student change. The bound also holds for mean
conditional JS: the joint distribution $P_X(s,v)=w_s p_X(v\mid s)$
uses the same prefix weights for all models, and its JS equals the
weighted mean of conditional JS.

\FloatBarrier
\subsection{Parameter and Prediction Distances}
\label{app:teacher_trajectories}

Figure~\ref{fig:self_opd_trajectory_comparison} presents Qwen3-4B using the
same student--teacher pairs for parameter and prediction measurements.
Table~\ref{tab:self_opd_endpoint_distances} reports the corresponding
step-780 distances for the other four models, using the same paired
students and teachers for both measurements.
Parameter distances are Euclidean distances across all
model parameters, normalized by the corresponding Base--teacher endpoint distance.
Let $d_T=\theta_T-\theta_0$ and $d_S(t)=\theta_{S_t}-\theta_0$ be the
teacher and student displacements. The normalized distances are
\begin{equation}
    d_\theta(t)=\frac{\|\theta_{S_t}-\theta_T\|_2}{\|d_T\|_2},
    \qquad
    d_\perp(t)=\frac{\|d_S(t)-\alpha_t d_T\|_2}{\|d_T\|_2},
    \quad
    \alpha_t=\frac{\langle d_S(t),d_T\rangle}{\|d_T\|_2^2}.
    \label{eq:parameter_distances}
\end{equation}
Here $\alpha_t$ projects the student displacement onto the teacher
direction; $d_\perp$ includes the residual in all orthogonal directions
and is zero for models on the Base--teacher line.
The teacher endpoint is fixed throughout each trajectory, including
the reference trajectory of the source training.

Prediction trajectories compare students of SFT and RLVR teachers with their
respective teachers on fixed evaluation prefixes.
In-domain tasks are AMC23, AIME24, AIME25, HMMT25, and Math500;
OOD tasks are HellaSwag, ARC-Challenge, MMLU, IFEval, TriviaQA, BBH,
and NQ. We average JS equally across datasets within each domain,
then divide by the initial Base--teacher JS on the same prefixes.
The normalized distance is one at Base. Each curve keeps its teacher
and prefixes fixed.

Across the five models, Self-OPD students remain far from their teachers
in parameter space and depart from the Base--teacher line as their
prediction gaps shrink. Students of RLVR teachers approach teacher
in-domain predictions closely; students of SFT teachers generally
retain larger in-domain gaps. Both sources reduce OOD prediction gaps.

\begin{table}[!htb]
    \centering
    \small
    \setlength{\tabcolsep}{7pt}
    \begin{tabular}{@{}llrrrr@{}}
\toprule
Student & Teacher & \multicolumn{2}{c}{Parameter distance} & \multicolumn{2}{c}{Prediction distance} \\
\cmidrule(lr){3-4}\cmidrule(l){5-6}
& & $d_\theta$ & $d_\perp$ & $d_{\mathrm{out,in}}$ & $d_{\mathrm{out,OOD}}$ \\
\midrule
Qwen2.5-1.5B & SFT & 0.825 & 0.355 & 0.493 & 0.361 \\
 & RLVR & 0.894 & 0.744 & 0.010 & 0.354 \\
\addlinespace[3pt]
Qwen2.5-7B & SFT & 0.976 & 0.471 & 0.601 & 0.173 \\
 & RLVR & 0.943 & 0.758 & 0.010 & 0.243 \\
\addlinespace[3pt]
Qwen3-1.7B & SFT & 1.003 & 0.516 & 0.705 & 0.187 \\
 & RLVR & 0.988 & 0.860 & 0.029 & 0.101 \\
\addlinespace[3pt]
Qwen3-8B & SFT & 0.998 & 0.457 & 0.084 & 0.238 \\
 & RLVR & 1.009 & 0.796 & 0.077 & 0.172 \\
\bottomrule
\end{tabular}

    \caption{Self-OPD distances to the corresponding SFT or RLVR teacher
    at step 780 for the four models supplementing
    Figure~\ref{fig:self_opd_trajectory_comparison}.
    Parameter and prediction distances are normalized by the initial
    Base--teacher gap; $d_\perp$ measures departure from the Base--teacher line.}
    \label{tab:self_opd_endpoint_distances}
\end{table}

\FloatBarrier
\subsection{Paired Capability Curves and Evaluation}
\label{app:rawraw_source_curves}

\begin{figure}[!htbp]
    \centering
    \includegraphics[width=\textwidth]{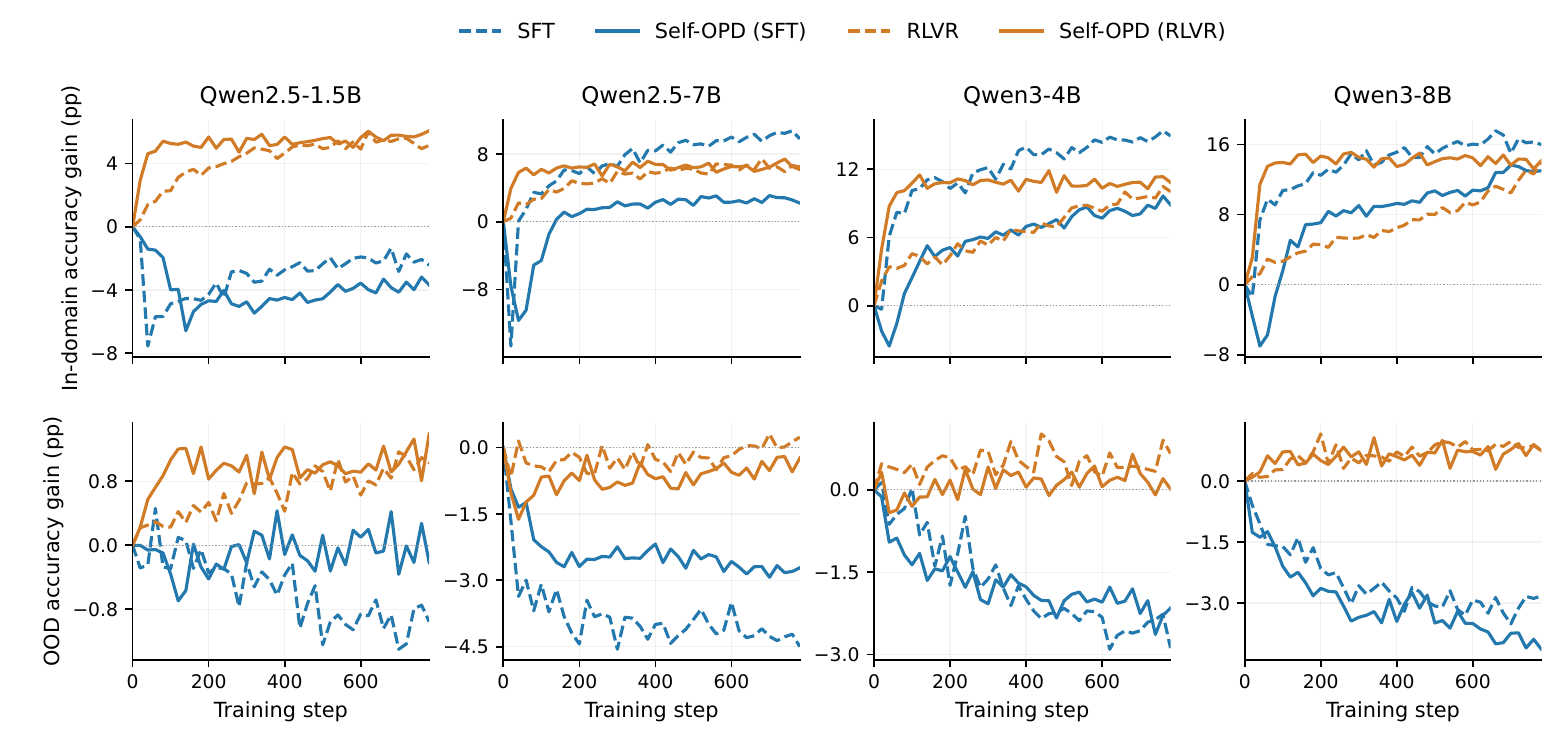}
    \caption{Accuracy changes from Base for source training (dashed) and
    the corresponding Self-OPD student (solid).
    Students use a fixed endpoint teacher.}
    \label{fig:source_training_accuracy}
\end{figure}

Each Self-OPD student is paired with the SFT or RLVR run that produced
its frozen teacher. Figure~\ref{fig:source_training_accuracy} uses completed
in-domain and seven-task OOD evaluations.

Source SFT/RLVR and Self-OPD evaluations use the same OOD scoring rules
and equal dataset weights.
RLVR students approach teacher task performance early, while larger
SFT students inherit substantial OOD losses before matching task gains.

\FloatBarrier
\subsection{Teacher-Prefix Probabilities Under the Student}
\label{app:teacher_prefix_probability}

We use Qwen2.5-7B and Qwen3-1.7B for this diagnostic. We fix 64 OpenR1
mathematical problems and generate four responses per problem from each
SFT and RLVR endpoint teacher, using Raw inputs,
temperature 1, top-$p=1$, and a maximum response length of 1024 tokens.
For the first $L$ tokens of a teacher response, Base, both Self-OPD
students, and the generating teacher each compute
\begin{equation}
    \ell_L(p;y\mid x)
    =-\frac{1}{L}\sum_{t=1}^{L}\log p(y_t\mid x,y_{<t}).
\end{equation}
Lower scores mean higher probability for the same fixed-length prefix.
We first average responses within a problem, then weight problems equally.

Table~\ref{tab:teacher_prefix_cross} gives cross-scoring results for two
models at the same student checkpoints, 60 and 240. Both students
increase the probability of their teachers' prefixes. RLVR prefixes
already have higher probability under Base, and their students approach
teacher self-scores faster.

\begin{table}[!htb]
    \centering
    \small
    \setlength{\tabcolsep}{4pt}
    \begin{tabular}{lrrrrrr}
\toprule
Prefix source & Base & SFT@60 & SFT@240 & RLVR@60 & RLVR@240 & Teacher \\
\midrule
\multicolumn{7}{l}{\textbf{Qwen2.5-7B}} \\
SFT & 1.057 & 0.871 & 0.580 & 1.563 & 1.567 & 0.379 \\
RLVR & 0.230 & 0.330 & 0.436 & 0.106 & 0.096 & 0.094 \\
\addlinespace
\multicolumn{7}{l}{\textbf{Qwen3-1.7B}} \\
SFT & 0.967 & 0.839 & 0.665 & 1.855 & 1.822 & 0.458 \\
RLVR & 0.300 & 0.332 & 0.444 & 0.145 & 0.136 & 0.135 \\
\addlinespace
\bottomrule
\end{tabular}

    \caption{Mean negative log probability over the first 128 tokens
    of teacher responses, in nats/token.
    SFT/RLVR columns denote students of the corresponding teacher source;
    @ denotes student training step.
    Each teacher contributes all 256 responses; the final column is
    scored by the teacher that generated them.}
    \label{tab:teacher_prefix_cross}
\end{table}

\paragraph{Correcting for teacher self-probability.}
Prefix scores also reflect how concentrated the teacher's own
generation distribution is. We therefore subtract teacher
self-scored negative log probability from the Base score:
\begin{equation}
    g_L(q,p_0;y\mid x)
    =\frac{1}{L}\sum_{t=1}^{L}
    \log\frac{q(y_t\mid x,y_{<t})}{p_0(y_t\mid x,y_{<t})}.
\end{equation}
Table~\ref{tab:teacher_prefix_correction} shows a larger teacher--Base
scoring gap for SFT responses.
This relationship persists when scoring only tokens 33--128,
extending the source difference to later parts of the response.

\begin{table}[!htb]
    \centering
    \small
    \begin{tabular}{lrr}
\toprule
Prefix source & Tokens 1--128 & Tokens 33--128 \\
\midrule
\multicolumn{3}{l}{\textbf{Qwen2.5-7B}} \\
SFT & 0.678 & 0.514 \\
RLVR & 0.136 & 0.122 \\
\addlinespace
\multicolumn{3}{l}{\textbf{Qwen3-1.7B}} \\
SFT & 0.509 & 0.353 \\
RLVR & 0.165 & 0.132 \\
\addlinespace
\bottomrule
\end{tabular}

    \caption{Scoring gaps after correcting for teacher self-probability,
    in nats/token. Larger values indicate a higher predictive cost
    under Base relative to the generating teacher.}
    \label{tab:teacher_prefix_correction}
\end{table}

\FloatBarrier

\subsection{SFT Teacher Openings and First-Position Feedback}
\label{app:sft_opening_feedback}

\paragraph{Opening preferences and candidate coverage.}
Of the 49,993 complete OpenR1 training solutions, 99.54\% begin with
\texttt{Okay} or \texttt{Alright} after stripping whitespace and ignoring
case. SFT retains these openings in its training targets.
On 64 fixed mathematical problems with four responses per problem,
all 256 responses from each of the Qwen2.5-7B and Qwen3-1.7B SFT
teachers begin with these words; none from their RLVR counterparts do.
At each response position, OPD computes feedback on the student's
16 highest-probability tokens. On the same Base-student responses,
all positive probability mass added by either SFT teacher relative to
Base at the first position falls outside this candidate set.
At later sampled positions, the excluded fractions are only 7.02\%
and 8.87\%, respectively, weighted by added probability mass.
Thus, the teacher assigns low probability to the student's usual
openings while its preferred openings receive no direct candidate-level
positive feedback.

\paragraph{Removing first-position feedback improves task transfer.}
We compare ordinary Self-OPD (SFT) with first-position feedback zeroed
throughout a complete 780-step run for each of the five Qwen models.
Each pair starts from the same pretrained Base and uses the same frozen
SFT teacher at its final checkpoint, with Raw inputs for both student
and teacher.
The intervention zeros all 16 candidate feedback values at the first
valid response position; later feedback, response masks, and the loss
denominator are unchanged. Training follows the standard Self-OPD
recipe with Top-16 feedback, AdamW, and 64 problems with four responses
per problem in each batch. Evaluation uses ordinary generation and
scoring, averaging datasets equally within the five-task in-domain
and seven-task OOD groups. The common target evaluation protocol is
specified in Appendix~\ref{app:experimental_details}.

\begin{table}[!htb]
    \centering
    \small
    \setlength{\tabcolsep}{4pt}
    \begin{tabular}{@{}lrrr@{\hspace{16pt}}rrr@{}}
\toprule
& \multicolumn{3}{c}{In-domain accuracy} & \multicolumn{3}{c}{OOD accuracy} \\
\cmidrule(lr){2-4}\cmidrule(l){5-7}
Model & Ordinary & Zeroed & $\Delta$ & Ordinary & Zeroed & $\Delta$ \\
\midrule
Qwen2.5-1.5B & 10.33 & 14.23 & +3.91 & 34.93 & 34.58 & -0.35 \\
Qwen2.5-7B & 28.49 & 34.57 & +6.08 & 47.02 & 46.50 & -0.52 \\
Qwen3-1.7B & 17.16 & 20.54 & +3.38 & 35.59 & 35.71 & +0.12 \\
Qwen3-4B & 34.71 & 38.82 & +4.11 & 43.45 & 43.35 & -0.10 \\
Qwen3-8B & 44.79 & 44.12 & -0.66 & 47.10 & 47.63 & +0.53 \\
\bottomrule
\end{tabular}

    \caption{Step-780 accuracy (\%) with ordinary feedback and
    first-position feedback zeroed. $\Delta$ is zeroed minus ordinary
    accuracy, in percentage points. Evaluation leaves all output
    positions unchanged.}
    \label{tab:sft_opening_intervention}
\end{table}

\begin{figure}[!htbp]
    \centering
    \includegraphics[width=\textwidth]{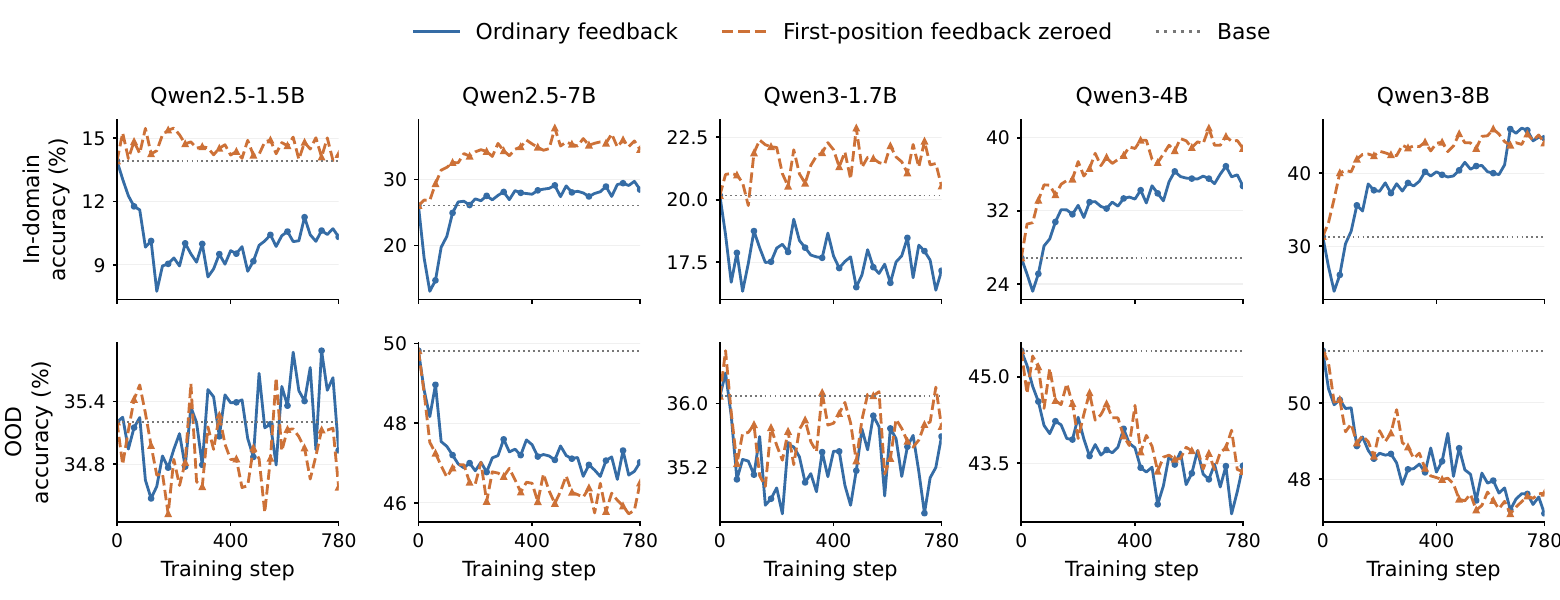}
    \caption{Complete accuracy trajectories for ordinary Self-OPD (SFT)
    and its first-position feedback intervention across five models
    (columns), with in-domain accuracy above OOD accuracy.
    Both conditions use the same Base and frozen SFT teacher.
    Dotted horizontal lines mark Base accuracy.}
    \label{fig:sft_first_position_accuracy}
\end{figure}

Zeroing first-position feedback generally improves in-domain accuracy,
with endpoint gains reaching 6.08 percentage points and OOD accuracy
differences within 0.53 percentage points
(Table~\ref{tab:sft_opening_intervention}).
Figure~\ref{fig:sft_first_position_accuracy} shows the complete training
curves. A separate short-run Qwen2.5-7B comparison with matched data order
and sampling seed also improves in-domain accuracy from 18.28\% to
27.92\% at step 20 and from 14.47\% to 27.53\% at step 40.
Together with the opening and coverage measurements, these results
show that first-position feedback can interfere with task transfer.

\paragraph{Task improvement and prediction matching.}
We also compare each student with Base and with its paired SFT teacher
on the same fixed evaluation prefixes. JS uses the equal mixture of
Base- and RLVR-generated prefixes, averaged equally over datasets
within each domain. Both comparisons use absolute JS, without
normalizing by the initial teacher--Base distance.
Accuracy is evaluated every 20 updates and JS every 60 updates.
Table~\ref{tab:sft_first_position_js} reports endpoint distances.

\begin{table}[!htb]
    \centering
    \small
    \setlength{\tabcolsep}{4pt}
    \begin{tabular}{@{}llrr@{\hspace{16pt}}rr@{}}
\toprule
& & \multicolumn{2}{c}{JS to Base} & \multicolumn{2}{c}{JS to SFT teacher} \\
\cmidrule(lr){3-4}\cmidrule(l){5-6}
Model & Feedback & In-domain & OOD & In-domain & OOD \\
\midrule
Qwen2.5-1.5B & Ordinary & 7.200 & 2.136 & 3.658 & 1.007 \\
 & First-position zeroed & 4.172 & 1.750 & 4.917 & 0.860 \\
\addlinespace[3pt]
Qwen2.5-7B & Ordinary & 7.137 & 1.963 & 4.625 & 0.462 \\
 & First-position zeroed & 5.355 & 1.620 & 4.779 & 0.753 \\
\addlinespace[3pt]
Qwen3-1.7B & Ordinary & 6.165 & 2.149 & 4.555 & 0.521 \\
 & First-position zeroed & 3.006 & 1.811 & 4.588 & 0.534 \\
\addlinespace[3pt]
Qwen3-4B & Ordinary & 6.941 & 2.122 & 4.026 & 0.688 \\
 & First-position zeroed & 5.880 & 2.071 & 3.023 & 0.548 \\
\addlinespace[3pt]
Qwen3-8B & Ordinary & 6.519 & 2.653 & 0.582 & 0.707 \\
 & First-position zeroed & 4.037 & 2.303 & 4.569 & 0.615 \\
\bottomrule
\end{tabular}

    \caption{Step-780 JS in $10^{-2}$ nats for the same ordinary and
    first-position-zeroed students. The references are the shared Base
    and the corresponding frozen SFT teacher.}
    \label{tab:sft_first_position_js}
\end{table}

Zeroing first-position feedback reduces both in-domain and OOD JS
from Base for all five models. The in-domain accuracy gains do not
consistently coincide with lower JS to the teacher. Improved task
performance can therefore accompany smaller changes from Base
without closer prediction matching to the teacher.

\FloatBarrier

\FloatBarrier

\subsection{Additional RLVR-Teacher Comparisons}
\label{app:rawraw_abc_accuracy}

Table~\ref{tab:teacher_source_abc_full} reports two additional teacher--student
pairs for Qwen3-1.7B at student step 120, using the A/D/E subset of conditions in
Table~\ref{tab:teacher_source_abc}. This fixed early-training comparison
uses complete in-domain accuracy, OOD accuracy, and JS measurements
for every group. The three student runs within each group use the same
initialization, update settings, and evaluation prefixes, with only the
teacher varying.

\begin{table}[!htb]
    \centering
    \small
    \setlength{\tabcolsep}{4pt}
    \begin{tabular}{@{}lrrrr@{\,}lr@{\,}l@{}}
\toprule
Teacher supervision & \multicolumn{3}{c}{Distribution change} & \multicolumn{4}{c}{Accuracy (\%)} \\
\cmidrule(lr){2-4}\cmidrule(l){5-8}
& $D_{\mathrm{in}}$ & $D_{\mathrm{OOD}}$ & $\rho$ & \multicolumn{2}{c}{Math ($\Delta$)} & \multicolumn{2}{c}{OOD ($\Delta$)} \\
\midrule
\multicolumn{8}{l}{\textbf{Qwen3-4B$\to$Qwen3-1.7B}} \\
Student Base & -- & -- & -- & 20.17 & \(\phantom{(+00.00)}\) & 36.09 & \(\phantom{(+00.00)}\) \\
A: Larger Base teacher & 0.2692 & 0.1772 & 0.658 & 19.38 & \((-0.78)\) & 35.96 & \((-0.13)\) \\
D: Larger RLVR teacher & 4.1255 & 0.3643 & 0.088 & 24.21 & \((+4.04)\) & 36.19 & \((+0.10)\) \\
E: RLVR increment & 4.1056 & 0.3302 & 0.080 & 24.08 & \((+3.92)\) & 36.05 & \((-0.04)\) \\
\midrule
\multicolumn{8}{l}{\textbf{Qwen3-8B$\to$Qwen3-1.7B}} \\
Student Base & -- & -- & -- & 20.17 & \(\phantom{(+00.00)}\) & 36.09 & \(\phantom{(+00.00)}\) \\
A: Larger Base teacher & 0.3272 & 0.2875 & 0.879 & 20.65 & \((+0.49)\) & 35.94 & \((-0.15)\) \\
D: Larger RLVR teacher & 4.2713 & 0.3747 & 0.088 & 26.72 & \((+6.55)\) & 37.04 & \((+0.94)\) \\
E: RLVR increment & 4.1650 & 0.3205 & 0.077 & 27.02 & \((+6.85)\) & 36.66 & \((+0.57)\) \\
\bottomrule
\end{tabular}

    \caption{Additional A/D/E comparisons for Qwen3-1.7B at student step 120. JS is relative to
    student Base in $10^{-2}$ nats. Accuracies are percentages, with
    changes from Base in parentheses. Domains use equal dataset weights.}
    \label{tab:teacher_source_abc_full}
\end{table}

\FloatBarrier
\subsection{Distribution Change and Forgetting: Additional Model}

\begin{figure}[!htbp]
    \centering
    \includegraphics[width=\textwidth]{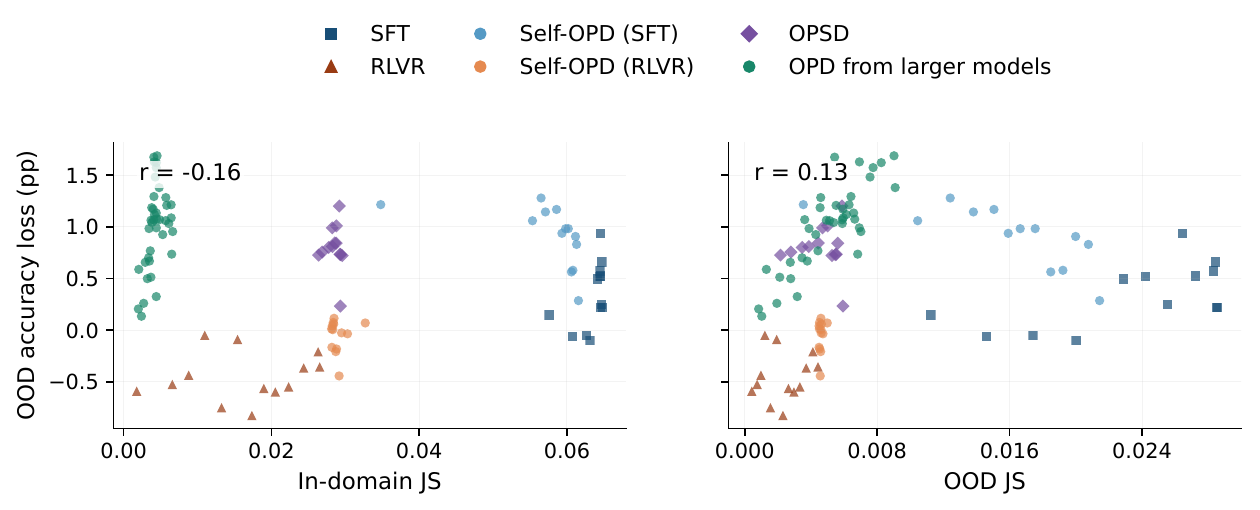}
    \caption{Distribution change and forgetting for Qwen3-1.7B.
    Left: in-domain JS; right: OOD JS. Both panels show OOD accuracy
    loss at the corresponding checkpoint, with $r$ denoting Pearson correlation.}
    \label{fig:distribution_forgetting_q3_1p7b}
\end{figure}

Figure~\ref{fig:distribution_forgetting_q3_1p7b} supplements
Figure~\ref{fig:distribution_forgetting} with Qwen3-1.7B,
using the same methods, evaluation prefixes, and dataset averages.

\FloatBarrier
\subsection{Training and Measurement Settings}
\label{app:rawraw_training_settings}

All A--E comparisons use student step 120 and the training recipe in
Appendix~\ref{app:training_settings}: Top-16 feedback, AdamW, and
64 problems with four responses per problem. Student and teacher
inputs are Raw. Within each comparison, initialization, training data,
sampling, and optimization are fixed while the teacher varies.
JS uses shared evaluation prefixes, with equal weights for Base- and
RLVR-generated prefixes and equal dataset weights within each domain.

\subsection{Constructed-Teacher Normalization and Current-Student Feedback}
\label{app:rawraw_source_controls}

Conditions C and E use $m\in\{\mathrm{SFT},\mathrm{RLVR}\}$ and
the constructed teacher $q_{\mathrm{inc},m}$ defined in
Section~\ref{sec:teacher_information}. The models within each construction
share a vocabulary. At each prefix, we combine logits from the frozen
smaller Base, larger Base, and larger post-trained model, writing $z_r$
for the logits of model $r$:
\begin{equation}
    q_{\mathrm{inc},m}
    =\operatorname{softmax}(z_{p_0}+z_{q_m}-z_{q_{\mathrm{base}}}).
\end{equation}
The resulting ordinary OPD feedback is
\begin{equation}
    \log\frac{q_{\mathrm{inc},m}(v\mid s)}{p_t(v\mid s)}
    =\delta_m(v\mid s)
      -\log\frac{p_t(v\mid s)}{p_0(v\mid s)}-\log Z_m(s).
\end{equation}
The student's current change from Base is subtracted as training
proceeds, while the teacher's source models remain frozen.

\FloatBarrier

\section{OPSD: Teacher Input and Answer Boundaries}
\label{app:opsd_formats}
\label{app:opsd_answer_boundary}

OPSD uses privileged reference solutions to provide token-level feedback
on student-generated answers. Its original implementation uses
chat-formatted teacher prompts, which provide an explicit transition
from reference material to the response being scored~\citep{zhao2026opsd}.
We adapt this transition to the Raw text-completion setting of our Base
models and compare three ways of presenting the same reference information.

\subsection{Three Teacher Input Formats}
\label{app:opsd_three_formats}

In all three settings, the student receives the Raw problem and generates
its own response. The teacher uses the same Base weights as the student's
initialization, receives the problem and full reference solution, and scores
the student-generated tokens. The private reference block and instruction
are given in Appendix~\ref{app:training_input_formats}.
Only the teacher's input formatting differs:
\begin{itemize}
    \item \textbf{Raw:} concatenate the problem, private reference block,
    and instruction as plain text, followed directly by the student response.
    \item \textbf{Chat template:} package the same material as a user message
    and place the template's assistant-start marker before the student response.
    \item \textbf{Raw + answer boundary:} retain the Raw input and append
    \texttt{Answer to the original problem:} on a new line before the response.
\end{itemize}
The resulting teacher prefixes have the following structure, where
\texttt{[private block]} includes the reference and its instruction:
\begingroup\footnotesize
\begin{verbatim}
Raw:       [problem] [private block] [student response]
Chat:      <user> [problem] [private block] </user>
           <assistant> [student response]
Boundary:  [problem] [private block]
           Answer to the original problem:
           [student response]
\end{verbatim}
\endgroup
The role markers above are schematic; each model uses its corresponding
template.
Teacher weights remain those of the Base model in every setting.
The original OPSD prompt likewise explicitly transitions from the
reference solution to a new answer~\citep{zhao2026opsd}.

\subsection{Preserving the Transition to a New Answer}

Holding 64 mathematical problems and student responses fixed, we compare
the three teacher prefixes before the first response token.
Table~\ref{tab:opsd_boundary_stop} reports the teacher's mean probability
of the end-of-sequence token at this position. Direct Raw concatenation
gives the teacher a strong preference to stop. Both the assistant-start
marker and the textual answer boundary sharply reduce that preference.

\begin{table}[!htb]
    \centering
    \small
    \begin{tabular}{lrrr}
        \toprule
        Teacher & Raw & Chat template & Raw + answer boundary \\
        \midrule
        Qwen2.5-7B & 24.12 & 0.0100 & 0.0028 \\
        Qwen3-1.7B & 15.36 & 0.0385 & 0.0034 \\
        \bottomrule
    \end{tabular}
    \caption{Teacher end-of-sequence probability (\%) before the first
    student response token, with the problems and reference solutions fixed.}
    \label{tab:opsd_boundary_stop}
\end{table}

The stopping preference also affects training. In paired short Qwen2.5-7B
runs, both settings begin with 12.11\% empty student responses in the
first batch. By the fourth batch, this rises to 21.09\% with direct Raw
concatenation and falls to 1.95\% with the textual boundary. Each batch
contains the same 64 problems with four responses per problem.
These measurements suggest that simply removing the chat template can
leave the Base teacher without a clear distinction between the completed
reference solution and the student's new answer. The assistant-start marker
in the original prompting convention supplies this distinction; an explicit
textual boundary preserves its role in a Raw prompt. Both forms of separation
help the teacher provide feedback on the student's own solution.

\subsection{Full Training Across Qwen Models}
\label{app:opsd_full_training}

\begin{figure}[p]
    \centering
    \includegraphics[width=\textwidth,height=.86\textheight,keepaspectratio]{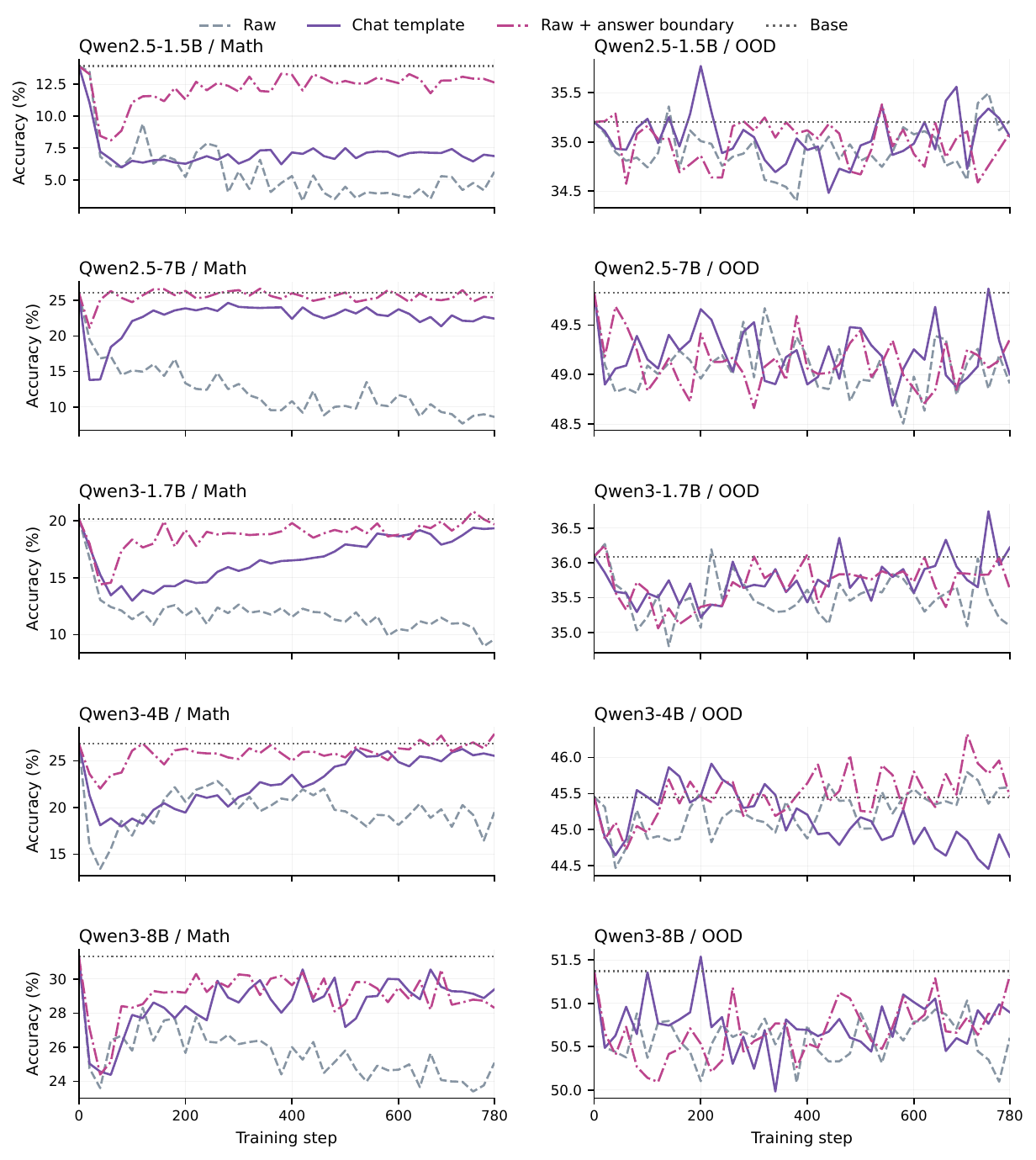}
    \caption{OPSD mathematical and OOD accuracy throughout training
    under the three teacher input formats, one student per row.
    Dotted horizontal lines mark the common Base evaluation. Each curve
    uses measured checkpoints at 20-step intervals.}
    \label{fig:opsd_format_curves}
\end{figure}

We compare complete 780-step runs for the five Qwen students.
Students use OpenR1-50K, Top-16 feedback, AdamW,
and the shared training settings in Appendix~\ref{app:experimental_details}.
Table~\ref{tab:opsd_format_accuracy} gives final mathematical and OOD
accuracy; Table~\ref{tab:opsd_format_js} gives the corresponding JS and
$\rho$. Figure~\ref{fig:opsd_format_curves} shows the accuracy trajectories.
Each entry uses one training seed. The chat-template entries retain the
original prompting convention, and the boundary entries are used in our
main accuracy comparisons. Mathematical and OOD scores
use the same five and seven datasets as the main evaluation, respectively.
Distribution measurements compare each student with its Base on shared
fixed prefixes, using the same task groups and averaging as
Figure~\ref{fig:predictive_distributions}.

\begin{table}[!htb]
    \centering
    \small
    \setlength{\tabcolsep}{5pt}
    \begin{tabular}{llrrrr}
\toprule
Student & Metric & Base & Raw & Chat template & Raw + boundary \\
\midrule
Qwen2.5-1.5B & Math & 13.93 & 5.63 & 6.86 & 12.64 \\
 & OOD & 35.20 & 35.21 & 35.06 & 35.09 \\
\addlinespace[3pt]
Qwen2.5-7B & Math & 26.08 & 8.61 & 22.44 & 25.46 \\
 & OOD & 49.82 & 48.91 & 49.00 & 49.35 \\
\addlinespace[3pt]
Qwen3-1.7B & Math & 20.16 & 9.62 & 19.34 & 19.68 \\
 & OOD & 36.09 & 35.09 & 36.22 & 35.63 \\
\addlinespace[3pt]
Qwen3-4B & Math & 26.85 & 19.51 & 25.55 & 27.89 \\
 & OOD & 45.45 & 45.59 & 44.62 & 45.46 \\
\addlinespace[3pt]
Qwen3-8B & Math & 31.32 & 25.10 & 29.40 & 28.31 \\
 & OOD & 51.37 & 50.60 & 50.90 & 51.32 \\
\bottomrule
\end{tabular}

    \caption{OPSD accuracy (\%) at step 780 under three teacher input
    formats. Students always receive Raw problems. Base is the untrained
    student; each metric averages its datasets equally.}
    \label{tab:opsd_format_accuracy}
\end{table}

\begin{table}[!htb]
    \centering
    \small
    \setlength{\tabcolsep}{7pt}
    \begin{tabular}{llrrr}
\toprule
Student & Teacher input & $D_{\mathrm{in}}$ & $D_{\mathrm{OOD}}$ & $\rho$ \\
\midrule
Qwen2.5-1.5B & Raw & 2.965 & 0.440 & 0.148 \\
 & Chat template & 4.978 & 0.472 & 0.095 \\
 & Raw + answer boundary & 4.691 & 0.599 & 0.128 \\
\addlinespace[3pt]
Qwen2.5-7B & Raw & 2.308 & 0.189 & 0.082 \\
 & Chat template & 4.789 & 0.192 & 0.040 \\
 & Raw + answer boundary & 4.585 & 0.188 & 0.041 \\
\addlinespace[3pt]
Qwen3-1.7B & Raw & 2.920 & 0.587 & 0.201 \\
 & Chat template & 4.904 & 0.571 & 0.117 \\
 & Raw + answer boundary & 4.648 & 0.706 & 0.152 \\
\addlinespace[3pt]
Qwen3-4B & Raw & 3.278 & 0.244 & 0.074 \\
 & Chat template & 4.495 & 0.318 & 0.071 \\
 & Raw + answer boundary & 4.488 & 0.317 & 0.071 \\
\addlinespace[3pt]
Qwen3-8B & Raw & 3.280 & 0.266 & 0.081 \\
 & Chat template & 4.416 & 0.185 & 0.042 \\
 & Raw + answer boundary & 4.549 & 0.295 & 0.065 \\
\bottomrule
\end{tabular}

    \caption{OPSD distribution changes at step 780 for the same runs as
    Table~\ref{tab:opsd_format_accuracy}. JS is measured relative to Base
    in $10^{-2}$ nats; $\rho=D_{\mathrm{OOD}}/D_{\mathrm{in}}$.}
    \label{tab:opsd_format_js}
\end{table}

Adding the textual answer boundary improves mathematical accuracy over
direct Raw concatenation for all five students.
It also exceeds the chat-template result for four of the five students.
Across formats,
OOD accuracy changes much less than mathematical accuracy.
Thus, separating the reference from the new answer is consequential for
learning, and the most effective realization of that boundary depends on
the model.

\paragraph{Configurations used in the main comparisons.}
The accuracy comparisons in Figure~\ref{fig:overview}(b) and
Table~\ref{tab:main_results} use a Raw student and a Raw teacher with the
explicit answer boundary. This preserves the reference-to-answer
transition in the original OPSD prompting convention while adapting it
to Base models' text-completion format. The distribution comparison in
Figure~\ref{fig:predictive_distributions} retains the plain Raw/Raw control
to match teacher and student formatting without additional answer-start
cues. Both configurations provide the same privileged reference information
to the teacher. The complete accuracy and JS comparisons here document
the effects of this input adaptation.

\paragraph{Teacher input can affect update selectivity.}
Interestingly, Chat formatting consistently raises in-domain JS relative
to Raw, while changes in OOD JS are much smaller in absolute size and
its five-model mean remains nearly unchanged. Consequently, $\rho$
decreases. A textual answer boundary produces in-domain JS close to
Chat formatting, showing that ordinary language cues can have a similar
effect to role markers. These results suggest that teacher conditioning
can introduce response preferences whose effects concentrate on the
target domain. Together with the boundary variant's accuracy improvements
over plain Raw, they motivate optimizing teacher prompts to direct OPD
feedback toward task learning while limiting accompanying OOD changes.

\clearpage
\clearpage

\clearpage
\section{Weight Interpolation and Extrapolation}
\label{app:weight_interpolation}
\label{app:rawraw_weight_scaling}
\begingroup
\setlength{\textfloatsep}{8pt}
\setlength{\intextsep}{8pt}

Section~\ref{sec:weight_scaling} compares SFT, RLVR, and their Self-OPD
students under weight scaling. We retain these four methods for all
five student models, using the original training seed and step-780
endpoints. Both the student and teacher use Raw input in Self-OPD.
The SFT/RLVR sweeps reuse the corresponding source-training endpoints.
The curves share the complete measured interval $0\leq\lambda\leq2$.
We rescale the weight difference between Base ($\theta_0$) and the trained
endpoint ($\theta_{\mathrm{end}}$), using the notation of Eq.~\eqref{eq:weight_scaling}:
\[
\theta_\lambda=\theta_0+\lambda(\theta_{\mathrm{end}}-\theta_0),
\qquad \lambda\in\{0,0.2,0.4,0.6,0.8,1,1.2,1.5,2\}.
\]
Values $0\leq\lambda\leq1$ interpolate between the endpoints;
$\lambda>1$ extrapolates beyond the trained endpoint. Evaluation follows
Appendix~\ref{app:experimental_details}. The points at $\lambda=0$ and
$\lambda=1$ use the evaluations of Base and the corresponding trained
endpoint, respectively.

\paragraph{Sweet-spot coefficients.}
Let $\Delta\mathrm{Acc}_{\mathrm{in}}(\lambda)$ and
$\Delta\mathrm{Acc}_{\mathrm{OOD}}(\lambda)$ be accuracy changes from Base,
in percentage points. We mark sweet spots only for runs whose trained
endpoint improves in-domain accuracy over Base,
$\Delta\mathrm{Acc}_{\mathrm{in}}(1)>0$.
Stars in Figure~\ref{fig:weight_scaling_accuracy} identify measured
coefficients satisfying
\[
    \lambda>0,
    \qquad
    \begin{aligned}
        \Delta\mathrm{Acc}_{\mathrm{in}}(\lambda)
        &>\Delta\mathrm{Acc}_{\mathrm{in}}(1),\\
        \Delta\mathrm{Acc}_{\mathrm{OOD}}(\lambda)
        &>\Delta\mathrm{Acc}_{\mathrm{OOD}}(1).
    \end{aligned}
\]
Both task means must be available at the candidate coefficient and
at $\lambda=1$. OOD accuracy is the equally weighted mean over the
same tasks as the main evaluation.

The identified sweet spots lie within the interpolation range for
SFT-based updates and the extrapolation range for RLVR-based updates.
Learning rates and training progress can also affect the endpoint
and the locations of these coefficients.

\paragraph{Learning and retention along a training direction.}
Interpolation toward Base can retain much of the in-domain gain while
recovering OOD accuracy (Figure~\ref{fig:weight_scaling_accuracy}).
Further extrapolation often reduces in-domain accuracy and accentuates
SFT-family forgetting. RLVR directions preserve OOD accuracy over a
broader range of weight scaling.

\clearpage
\endgroup

\clearpage
\section{Supporting Analysis}
\label{app:theory}

\subsection{Student Probabilities in the Expected OPD Gradient}
\label{app:rawraw_target_theory}
\label{app:local_score_updates}

At a fixed prefix $s$, let $p=\operatorname{softmax}(z)$ and sample
$a\sim p$. For the score-function surrogate
$L=-\operatorname{sg}(A)\log p(a)$, where $\operatorname{sg}$ denotes
stop-gradient, define $m_v=\mathbb E[A\mid a=v,s]$.
The gradient with respect to logit $z_v$ and its conditional mean are
\begin{align}
    g_v=\frac{\partial L}{\partial z_v}
       &=-A\bigl(\mathbf 1\{a=v\}-p(v)\bigr),\\
    \mu_v(s)=\mathbb E[g_v\mid s]
    &=-p(v)m_v+p(v)\sum_u p(u)m_u
      =p(v)\bigl(\mathbb E[A\mid s]-m_v\bigr).
    \label{eq:rawraw_mean_gradient}
\end{align}
For ordinary OPD, $m_v=\log q_T(v\mid s)-\log p(v\mid s)$.
The student's current token probabilities weight differences in teacher
preferences across candidates.
Applying the chain rule and averaging over student-generated prefixes
$d_p$ gives
\begin{equation}
    \mathbb E[\nabla_\theta L]
    =\mathbb E_{s\sim d_p}\left[\sum_v\mu_v(s)\nabla_\theta z_v(s)\right].
\end{equation}
Less frequently visited prefixes and lower-probability candidates receive
smaller weights in this mean update.
This full-vocabulary score-function model motivates the prefix analysis;
the implemented candidate feedback and optimizer are specified in
Appendix~\ref{app:rawraw_training_settings}.

Token-dependent feedback means that $m_v$ varies with candidate $v$
at a fixed prefix. Random per-token feedback with the same conditional
mean for every candidate still yields zero mean logit gradient.
As student predictions approach the frozen teacher, ordinary OPD's
log-probability feedback and its mean gradient shrink.

\subsection{Local Directions After Calibrating In-Domain Change}
\label{app:rawraw_matched_directions}
\label{app:local_influence}

\paragraph{A directional view of local JS.}
Let $p_0=p_{\theta_0}$ be student Base and consider a small update
$\theta_\epsilon=\theta_0+\epsilon g$ along a fixed parameter direction $g$.
For domain $d\in\{\mathrm{in},\mathrm{OOD}\}$, fix evaluation prefixes
and their task weights, represented by $\mathcal S_d$.
Define the score vector at Base and the domain's predictive Fisher matrix:
\begin{equation}
    \psi_s(v)=
    \left.\nabla_\theta\log p_\theta(v\mid s)\right|_{\theta=\theta_0},
    \qquad
    \mathcal I_d=\mathbb E_{s\sim\mathcal S_d}
    \left[\sum_v p_0(v\mid s)\psi_s(v)\psi_s(v)^\top\right].
\end{equation}
$\mathcal I_d$ captures the local sensitivity of that domain's predictions to
parameter changes.
Using natural logarithms, smooth conditional distributions, and
positive probabilities, mean JS satisfies
\begin{equation}
    D_d(\theta_0,\theta_\epsilon)
    =\frac{\epsilon^2}{8}g^\top\mathcal I_d g+\mathcal O(\epsilon^3).
    \label{eq:rawraw_local_js}
\end{equation}
The per-prefix quadratic term of JS is
$\frac18\sum_v(p_{\theta_\epsilon}-p_0)^2/p_0$.
Substituting
$p_{\theta_\epsilon}(v\mid s)-p_0(v\mid s)
=\epsilon p_0(v\mid s)\psi_s(v)^\top g+\mathcal O(\epsilon^2)$
and averaging over fixed prefixes yields the result.

When $g^\top\mathcal I_{\mathrm{in}}g>0$,
\begin{equation}
    \rho(\epsilon,g)
    =\frac{D_{\mathrm{OOD}}(\theta_0,\theta_\epsilon)}
    {D_{\mathrm{in}}(\theta_0,\theta_\epsilon)}
    =\frac{g^\top\mathcal I_{\mathrm{OOD}}g}{g^\top\mathcal I_{\mathrm{in}}g}
      +\mathcal O(\epsilon).
    \label{eq:rawraw_local_rho}
\end{equation}
Update magnitude cancels in the ratio of quadratic terms, leaving
the relative influence of the direction on the two domains.
This motivates calibrating in-domain change to compare the OOD
influence of directions induced by different feedback sources.

\paragraph{Comparing calibrated directions.}
The local comparison reuses the same prefixes generated by Base
and computes A/D/E update directions separately.
For each direction, we adjust only its overall magnitude to reach
a common in-domain JS target on calibration prefixes.
We then measure in-domain JS, OOD JS, and $\rho$ on independent
evaluation prefixes, so the reported in-domain JS can differ.
Reference RLVR directions use parameter differences along the
student's original RLVR training, with the same calibration procedure.

\begin{table}[!htb]
    \centering
    \small
    \setlength{\tabcolsep}{4pt}
    \begin{tabular}{lrrr}
\toprule
Update direction & In-domain JS $\times 10^5$ & OOD JS $\times 10^5$ & $\rho$ \\
\midrule
\multicolumn{4}{l}{\textbf{Qwen3-1.7B}} \\
A: Base teacher & 1.079 & 0.659 & 0.611 \\
D: RLVR teacher & 1.072 & 0.079 & 0.074 \\
E: RLVR increment & 0.938 & 0.053 & 0.056 \\
RLVR parameter change (step 60) & 1.228 & 0.071 & 0.058 \\
RLVR parameter change (step 780) & 1.098 & 0.069 & 0.063 \\
\midrule
\multicolumn{4}{l}{\textbf{Qwen3-4B}} \\
A: Base teacher & 0.835 & 0.070 & 0.084 \\
D: RLVR teacher & 1.538 & 0.025 & 0.016 \\
E: RLVR increment & 1.573 & 0.024 & 0.015 \\
\bottomrule
\end{tabular}

    \caption{Independent-prefix measurements after calibrating in-domain
    JS to $10^{-5}$. Incorporating the teacher's RLVR changes reduces
    relative OOD influence and brings it closer to the paired direct
    RLVR parameter directions.}
    \label{tab:rawraw_matched_directions}
\end{table}

\FloatBarrier

\subsection{Teacher Information Relative to Base}
\label{app:teacher_information}

Fix a student prefix $s$ and let $s^{\mathrm T}$ contain the same
problem and continuation under the teacher's input format, including
privileged context when present. On a common finite vocabulary with
positive probabilities, the Base-relative teacher preference is
\begin{equation}
    \begin{aligned}
        b_q(v,s)
        &=\log\frac{q(v\mid s^{\mathrm T})}{p_0(v\mid s)}\\
        &=\log\frac{q(v\mid s^{\mathrm T})}
                        {p_0(v\mid s^{\mathrm T})}
          +\log\frac{p_0(v\mid s^{\mathrm T})}{p_0(v\mid s)}.
    \end{aligned}
    \label{eq:teacher_context_decomposition}
\end{equation}
The first term compares teacher and Base under identical inputs; the
second isolates the effect of the teacher input on the Base model.
For Self-OPD (SFT) and Self-OPD (RLVR), the first term contains the
changes acquired during teacher training. For a larger Base teacher, it
contains differences between pretrained models; a post-trained larger
teacher additionally includes preferences acquired during post-training. For OPSD, the teacher
is a frozen copy of Base, so the first term is zero and the signal
comes from the augmented input.

\paragraph{Information in the A--E teachers.}
For $m\in\{\mathrm{SFT},\mathrm{RLVR}\}$, $q_m$ denotes the full
post-trained teacher in condition B or D, and $q_{\mathrm{inc},m}$ the
constructed teacher in condition C or E. The shared-input decomposition
in Eq.~\eqref{eq:teacher_source_decomposition} gives
\begin{equation}
    \begin{aligned}
        b_{q_{\mathrm{base}}}&=\delta_{\mathrm{pre}},\\
        b_{q_m}&=\delta_{\mathrm{pre}}+\delta_m,\\
        b_{q_{\mathrm{inc},m}}&=\delta_m-\log Z_m(s),
    \end{aligned}
    \label{eq:constructed_base_preference}
\end{equation}
where $Z_m(s)=\sum_v p_0(v\mid s)\exp[\delta_m(v\mid s)]$.
The last equality follows directly from normalizing
$p_0\exp(\delta_m)$. It identifies the information retained
by the constructed teacher while accounting for its normalization.

Adding and subtracting $\log p_0(v\mid s)$ gives
\begin{equation}
    \log\frac{q(v\mid s^{\mathrm T})}{p_\theta(v\mid s)}
    =b_q(v,s)-\log\frac{p_\theta(v\mid s)}{p_0(v\mid s)}.
    \label{eq:teacher_base_preferences}
\end{equation}
At the initial student,
the entire coefficient is $b_q(v,s)$. As the student changes, its
Base-relative log probability is subtracted from this fixed teacher
preference. The per-token identity also holds on the student's
Top-16 support, before applying the detached weights used in training.
Student-generated prefixes and candidate weights determine which
teacher preferences contribute to each update.

For the full-vocabulary local objective, the same decomposition gives
\begin{equation}
    \begin{aligned}
        &\mathrm{KL}\bigl(p_\theta(\cdot\mid s)
                         \,\|\,q(\cdot\mid s^{\mathrm T})\bigr)\\
        &\quad=\mathrm{KL}\bigl(p_\theta(\cdot\mid s)
                               \,\|\,p_0(\cdot\mid s)\bigr)
        -\mathbb E_{v\sim p_\theta(\cdot\mid s)}[b_q(v,s)].
    \end{aligned}
    \label{eq:base_relative_kl}
\end{equation}
This follows by expanding the log ratio inside the expectation.
All teacher configurations share this form; their information
sources determine the additional preferences $b_q$. For conditions C and E,
\begin{equation}
    \mathrm{KL}(p_\theta\|q_{\mathrm{inc},m})
    =\mathrm{KL}(p_\theta\|p_0)
     -\mathbb E_{p_\theta}[\delta_m]+\log Z_m(s).
    \label{eq:constructed_teacher_kl}
\end{equation}
At a fixed prefix, $Z_m(s)$ is independent of student parameters, so its
gradient is zero. Both conditions therefore combine fitting the respective
post-training increment with a penalty for the current student's departure
from Base. On any candidate set, centering with common normalized weights
also cancels $\log Z_m(s)$.

\paragraph{Teacher distance and student change.}
For fixed evaluation prefixes, define the teacher's relative distance
from student Base by
\begin{equation}
    \rho_{\mathrm T}(q;p_0)
    =\frac{D_{\mathrm{OOD}}(q,p_0)}{D_{\mathrm{in}}(q,p_0)},
    \qquad D_{\mathrm{in}}(q,p_0)>0.
\end{equation}
The ratio compares the frozen target with student Base. The student's
$\rho$ instead measures changes reached along its optimization path,
whose directions also depend on student prefix probabilities and
logit Jacobians (Appendix~\ref{app:rawraw_target_theory}).

A fixed-Base signal compares teacher and Base on the same input:
$A_\Delta(v,s)=\log[q(v\mid s^{\mathrm T})/p_0(v\mid s^{\mathrm T})]$.
Using $p_0(\cdot\mid s^{\mathrm T})$ as the reference gives
\begin{equation}
    \begin{aligned}
        &\mathrm{KL}\bigl(p_\theta(\cdot\mid s)
                     \,\|\,q(\cdot\mid s^{\mathrm T})\bigr)\\
        &\quad=\mathrm{KL}\bigl(p_\theta(\cdot\mid s)
                     \,\|\,p_0(\cdot\mid s^{\mathrm T})\bigr)
          -\mathbb E_{v\sim p_\theta(\cdot\mid s)}[A_\Delta(v,s)].
    \end{aligned}
    \label{eq:delta_reference_kl}
\end{equation}
Thus the current-student reference balances the teacher--Base
preference against departure from the frozen reference distribution.

\subsection{Fixed-Base and Current-Student Feedback}
\label{app:delta_self_opd_signal}
\label{app:delta_reference_dynamics}

To examine the role of the current student in feedback, we compare
ordinary OPD with a fixed teacher-to-Base log ratio~\citep{heo2026opd2}.
Let $q$ be a frozen teacher, $p_0$ the frozen student Base, and
$p_\theta$ the current student. At a student-generated prefix,
$s$ denotes the student input and $s^{\mathrm T}$ includes the
teacher's conditioning information. Define
\begin{equation}
    \begin{aligned}
        A_{\mathrm{self}}(v,s)&=\log q(v\mid s^{\mathrm T})
            -\log p_\theta(v\mid s),\\
        A_\Delta(v,s)&=\log q(v\mid s^{\mathrm T})
            -\log p_0(v\mid s^{\mathrm T}).
    \end{aligned}
    \label{eq:delta_raw_signals}
\end{equation}
The fixed-Base signal compares teacher and Base under identical inputs.
All log probabilities use the full vocabulary and the same scoring
temperature. Conditions C and E use ordinary OPD feedback with
$q=q_{\mathrm{inc},m}$, so they also subtract the current student's
log probability (Appendix~\ref{app:rawraw_source_controls}).

\paragraph{Sampling and feedback both depend on the student.}
Let $\mu_\theta$ be the distribution of prefixes generated by the
current student. Compare the two raw feedback choices on the same prefixes drawn from
$\mu_\theta$. Ordinary OPD additionally uses $p_\theta$ in its
log-probability ratio (Eq.~\eqref{eq:delta_raw_signals}).
For full-vocabulary score updates with positive probabilities, define
$g_A=\mathbb E_{s\sim\mu_\theta,\,v\sim p_\theta(\cdot\mid s)}
[A(v,s)\nabla_\theta\log p_\theta(v\mid s)]$.
Holding sampled prefixes fixed when differentiating gives
\begin{equation}
    g_{\Delta}-g_{\mathrm{self}}
    =\mathbb E_{s\sim\mu_\theta}
      \left[\nabla_\theta\mathrm{KL}
      \bigl(p_\theta(\cdot\mid s)\|p_0(\cdot\mid s^{\mathrm T})\bigr)\right].
    \label{eq:delta_policy_correction}
\end{equation}
This follows by substituting the advantage difference and using
$\mathbb E_{p_\theta}\nabla_\theta\log p_\theta=0$.
The ordinary reference therefore corrects for the evolving student's
departure from the frozen Base distribution on the prefixes that
this same student visits.

\paragraph{Feedback decreases as preferences are acquired.}
Fix a prefix, use identical input formats, and suppress the prefix
notation. The teacher $q$ may also be a constructed $q_{\mathrm{inc},m}$.
Let $p_0$ and $q$ have positive probabilities on a finite
vocabulary, and consider the family
\begin{equation}
    b_v=\log\frac{q_v}{p_{0,v}},\qquad
    p_{\alpha,v}
    =\frac{p_{0,v}\exp(\alpha b_v)}{Z_\alpha},\qquad
    Z_\alpha=\sum_v p_{0,v}\exp(\alpha b_v),
    \quad 0\leq\alpha\leq1.
    \label{eq:delta_preference_path}
\end{equation}
The endpoints are $p_0$ and $q$. Substitution yields
\begin{equation}
    A_{\mathrm{self},v}=(1-\alpha)b_v+\log Z_\alpha,
    \qquad
    \overline A_{\mathrm{self},v}
    =(1-\alpha)\overline A_{\Delta,v},
    \label{eq:delta_attenuation_proof}
\end{equation}
where bars subtract the expectation under $p_\alpha$. The centered
identity also holds on a candidate set when both signals use the same
normalized centering weights.
The common constant has zero full-vocabulary score gradient.
Thus the current-student reference reduces feedback in proportion
to the remaining preference difference along this family.
At $\alpha=1$, the teacher--student signal and its mean update vanish.

\paragraph{A reward-aligned model of the RLVR source.}
\label{app:delta_reward_source}
For a fixed problem and a common input format, let $r(y)$ be the Base
distribution over a finite set of complete responses, with positive probabilities, and let
$R(y)\in\{0,1\}$ denote correctness. Consider a teacher that reweights
these responses by reward and a student along the same family:
\begin{equation}
    q_\beta(y)=\frac{r(y)e^{\beta R(y)}}{Z_\beta},
    \qquad
    p_\tau(y)=\frac{r(y)e^{\tau R(y)}}{Z_\tau},
    \qquad 0\leq\tau<\beta.
    \label{eq:delta_reward_tilt}
\end{equation}
The teacher is the optimizer of
$\mathbb E_q R-\beta^{-1}\mathrm{KL}(q\|r)$, as follows by
rewriting this objective as
$\beta^{-1}\log Z_\beta-\beta^{-1}\mathrm{KL}(q\|q_\beta)$.
This model represents training that changes the probability of
correctness while preserving Base's relative preferences among
responses with the same reward.

The token-level implication follows from conditioning on prefixes.
Let $V(s)=\Pr_r(R=1\mid s)$ and
$Z_b(s)=\mathbb E_r[e^{bR}\mid s]
=1+(e^b-1)V(s)$.
Since $q_b(v\mid s)=r(v\mid s)Z_b(sv)/Z_b(s)$, the two raw signals are
\begin{equation}
    \begin{aligned}
        A_\Delta(v,s)
        &=\log Z_\beta(sv)-\log Z_\beta(s),\\
        A_{\mathrm{self}}(v,s)
        &=\log\frac{Z_\beta(sv)}{Z_\tau(sv)}
          -\log\frac{Z_\beta(s)}{Z_\tau(s)}.
    \end{aligned}
    \label{eq:delta_reward_token_signals}
\end{equation}
For a fixed $s$, both signals are increasing functions of $V(sv)$.
Writing $k_b=e^b-1$, their derivatives with respect to $V(sv)$ are
\begin{equation}
    \frac{k_\beta}{1+k_\beta V(sv)}>0,
    \qquad
    \frac{k_\beta-k_\tau}
    {[1+k_\beta V(sv)][1+k_\tau V(sv)]}>0.
    \label{eq:delta_reward_preference_order}
\end{equation}
Thus the two references preserve the same ordering of token
continuations by eventual success in this model, with feedback
strength depending on the current student's progress.
The student's success probability,
$e^\tau V(sv)/Z_\tau(sv)$, is also increasing in $V(sv)$, so this
ordering agrees with expected correctness under the current student.
The statement holds at every visited prefix even as on-policy
rollouts change.

An SFT teacher fits a demonstration distribution whose probability
ratios relative to Base can also vary among equally correct responses.
Its fixed source signal can consequently retain response preferences
beyond correctness as the student learns. Eq.~\eqref{eq:delta_attenuation_proof} describes how the
current-student reference reduces the remaining feedback as these
preferences are acquired.
The effect of these preferences on retained capabilities depends on
their cross-domain influence, described in
Appendix~\ref{app:local_influence}.

\end{document}